\documentclass[10pt,twocolumn,letterpaper]{article}

\usepackage{wacv}
\usepackage{times}
\usepackage{epsfig}
\usepackage{graphicx}
\usepackage{amsmath}
\usepackage{amssymb}
\usepackage{booktabs}
\usepackage{caption}
\usepackage{array}
\usepackage{tabularx}
\usepackage[dvipsnames]{xcolor}
\usepackage[skip=0cm,list=true]{subcaption}
\usepackage{url}
\usepackage{multicol}
\usepackage{makecell}
\usepackage{comment}
\usepackage[accsupp]{axessibility}

\usepackage{cite}

\usepackage{microtype}
\usepackage{colortbl}

\usepackage{soul}
\usepackage{todonotes}

{\vspace{-\topsep}\begin{itemize}\itemsep1pt \parskip0pt \parsep0pt}
{\end{itemize}\vspace{-\topsep}}

\definecolor{gold(bg)}{rgb}{0.87, 0.79, 0.32}
\definecolor{gold(metallic)}{rgb}{0.83, 0.69, 0.22}
\definecolor{gold(web)(golden)}{rgb}{1.0, 0.84, 0.0}

\definecolor{palesilver}{rgb}{0.79, 0.75, 0.73}
\definecolor{silver}{rgb}{0.75, 0.75, 0.75}
\definecolor{lightslategray}{rgb}{0.47, 0.53, 0.6}

\definecolor{bronze(bg)}{rgb}{0.9, 0.6, 0.3}
\definecolor{bronze}{rgb}{0.8, 0.5, 0.2}

\definecolor{goldL}{HTML}{FBF2D2}
\definecolor{silverL}{HTML}{DDDDDD}
\definecolor{bronzeL}{HTML}{EED2B8}

\definecolor{goldD}{HTML}{D9AE13}
\definecolor{silverD}{HTML}{909090}
\definecolor{bronzeD}{HTML}{9A5F26}

\newcommand*\circledd[3]{\tikz[baseline=(char.base)]{
            \node[shape=circle,fill=#1,draw=#2,inner sep=1pt] (char) {\small{#3}};}}

\newcommand{\mfirst}[1]{%
    {#1\raisebox{0.8pt}{\circledd{goldL}{goldD}{1}}}%
}
\newcommand{\msecond}[1]{%
    {#1\raisebox{0.8pt}{\circledd{silverL}{silverD}{2}}}%
}
\newcommand{\mthird}[1]{%
    {#1\raisebox{0.8pt}{\circledd{bronzeL}{bronzeD}{3}}}%
}

\newcommand{\medal}[3]{\tikz[baseline=(char.base)]{\node[rounded corners=2pt,fill=#1,draw=#2,inner sep=1.5pt] (char) {#3};}}

\newcommand{\bm}[2]{
    \ifcase#1\or
      {\medal{goldL}{goldD}{\textbf{#2}}}
    \or 
      {\medal{silverL}{silverD}{#2}}
    \or 
      {\medal{bronzeL}{bronzeD}{#2}}
    \else 
      #2
    \fi\ignorespaces
}

\newcolumntype{L}[1]{>{\raggedright\let\newline\\\arraybackslash\hspace{0pt}}b{#1}}
\newcolumntype{C}[1]{>{\centering\let\newline\\\arraybackslash\hspace{0pt}}b{#1}}

\newcommand{\rankn}[1]{({\small\##1})}

\def\wacvPaperID{****} 

\wacvfinalcopy 

\ifwacvfinal
\usepackage[breaklinks=true,bookmarks=false,hidelinks=True]{hyperref}
\else
\usepackage[pagebackref=true,breaklinks=true,colorlinks,bookmarks=false]{hyperref}
\fi

\begin{document}

\title{2\textsuperscript{nd} Workshop on Maritime Computer Vision (MaCVi) 2024: Challenge Results}

\author{Benjamin Kiefer$^1$, Lojze Žust$^2$, Matej Kristan$^2$, Janez Perš$^2$, 
Matija Teršek$^3$, Arnold Wiliem$^{4,5}$, \\
Martin Messmer$^1$, Cheng-Yen Yang$^6$, Hsiang-Wei Huang$^6$, Zhongyu Jiang$^6$, Heng-Cheng Kuo$^6$,\\
Jie Mei$^6$, Jenq-Neng Hwang$^6$, Daniel Stadler$^{7,15}$, Lars Sommer$^{7,15}$, Kaer Huang$^8$,\\Aiguo Zheng$^9$, Weitu Chong$^{10}$, Kanokphan Lertniphonphan$^8$, Jun Xie$^8$, Feng Chen$^8$, Jian Li$^9$,\\
Zhepeng Wang$^8$, Luca Zedda$^{11}$, Andrea Loddo$^{11}$, Cecilia Di Ruberto$^{11}$, Tuan-Anh Vu$^{12}$,\\
Hai Nguyen-Truong$^{12}$, Tan-Sang Ha$^{12}$, Quan-Dung Pham$^{12}$, Sai-Kit Yeung$^{12}$, Yuan Feng$^{13}$,\\
Nguyen Thanh Thien$^{14}$, Lixin Tian$^{13}$, Sheng-Yao Kuan$^{6}$, Yuan-Hao Ho$^{6}$, Angel Bueno Rodriguez$^{16}$, \\
Borja Carrillo-Perez$^{16}$, Alexander Klein$^{16}$, Antje Alex$^{16}$, Yannik Steiniger$^{16}$, Felix Sattler$^{16}$, \\
Edgardo Solano-Carrillo$^{16}$,
Matej Fabijanić$^{17}$, Magdalena Šumunec$^{17}$,
Nadir Kapetanović$^{17}$, \\
Andreas Michel$^{7}$, Wolfgang Gross$^{7}$, Martin Weinmann$^{18}$
{\tt\small  }
\and
$^1$University of Tuebingen, 
$^2$University of Ljubljana, 
$^3$Luxonis,
$^{4}$Sentient Vision Systems,\\
$^{5}$Queensland University of Technology,
$^{6}$University of Washington,
$^{7}$Fraunhofer IOSB,\\
$^{8}$Lenovo Research,
$^{9}$Lenovo,
$^{10}$Fudan University,
$^{11}$University of Cagliari,\\
$^{12}$The Hong Kong University of Science and Technology,
$^{13}$Dalian Maritime University,\\
$^{14}$University of Information Technology,
$^{15}$Fraunhofer Center for Machine Learning,\\
$^{16}$German Aerospace Center,
$^{17}$University of Zagreb,
$^{18}$Karlsruhe Institute of Technology
}
\maketitle
\thispagestyle{empty}

\begin{abstract}
\vspace{0mm}
The 2$^{\text{nd}}$ Workshop on Maritime Computer Vision (MaCVi) 2024 
addresses maritime computer vision for Unmanned Aerial Vehicles (UAV) and Unmanned Surface Vehicles (USV). 
Three challenges categories are considered: (i) UAV-based Maritime Object Tracking with Re-identification, (ii) USV-based Maritime Obstacle Segmentation and Detection, (iii) USV-based Maritime Boat Tracking.
The USV-based Maritime Obstacle Segmentation and Detection features three sub-challenges, including a new embedded challenge addressing efficicent inference on real-world embedded devices.
This report offers a comprehensive overview of the findings from the challenges. We provide both statistical and qualitative analyses, evaluating trends from over 195 submissions. All datasets, evaluation code, and the leaderboard are available to the public at \url{https://macvi.org/workshop/macvi24}.
\end{abstract}

\section{Introduction}

Maritime environments, encompassing both the vast open seas and intricate coastlines, have always been of paramount importance for global trade, exploration, and scientific research. Over the past years, there has been a rapid increase in the deployment of autonomous robotic platforms, particularly Unmanned Aerial Vehicles (UAVs) and Unmanned Surface Vehicles (USVs), to augment and sometimes even replace traditional human-driven operations in this domain. These technological marvels, while offering unparalleled advantages in terms of scalability, efficiency, and safety, heavily rely on state-of-the-art computer vision systems to navigate, detect, and interpret their maritime surroundings.

\begin{figure}[t]
\centering

\begin{subfigure}{.45\textwidth}
   \includegraphics[width=\textwidth]{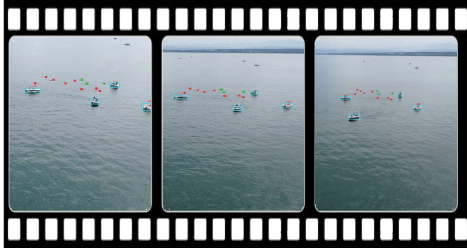}
   \caption{UAV-based Maritime MOT w/ re-identification}
\label{fig:g2}
\end{subfigure}
\begin{subfigure}{.45\textwidth}
   \includegraphics[width=\textwidth]{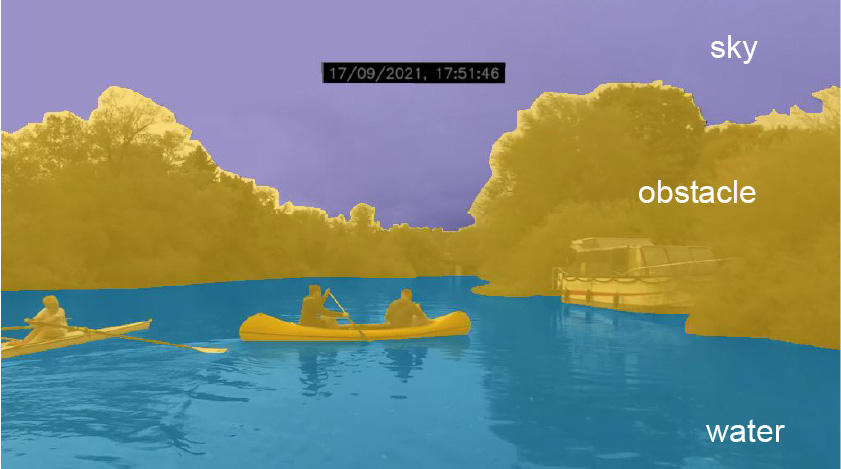}
   \captionsetup{justification=centering}
   \caption{USV-based Obstacle Segmentation \& \newline USV-based Embedded Obstacle Segmentation}
\label{fig:g3}
\end{subfigure}
\begin{subfigure}{.45\textwidth}
   \includegraphics[width=\textwidth]{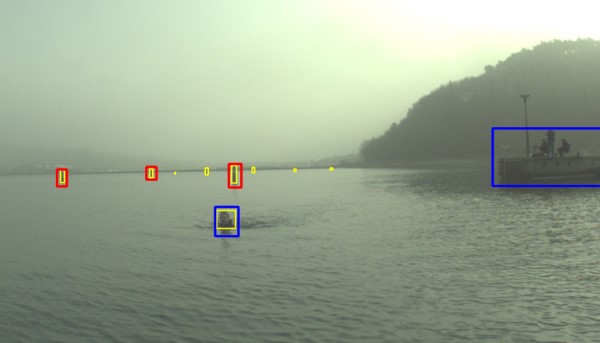}
   \caption{USV-based Maritime Obstacle Detection}
\label{fig:g4}
\end{subfigure}
\begin{subfigure}{.45\textwidth}
   \includegraphics[width=\textwidth]{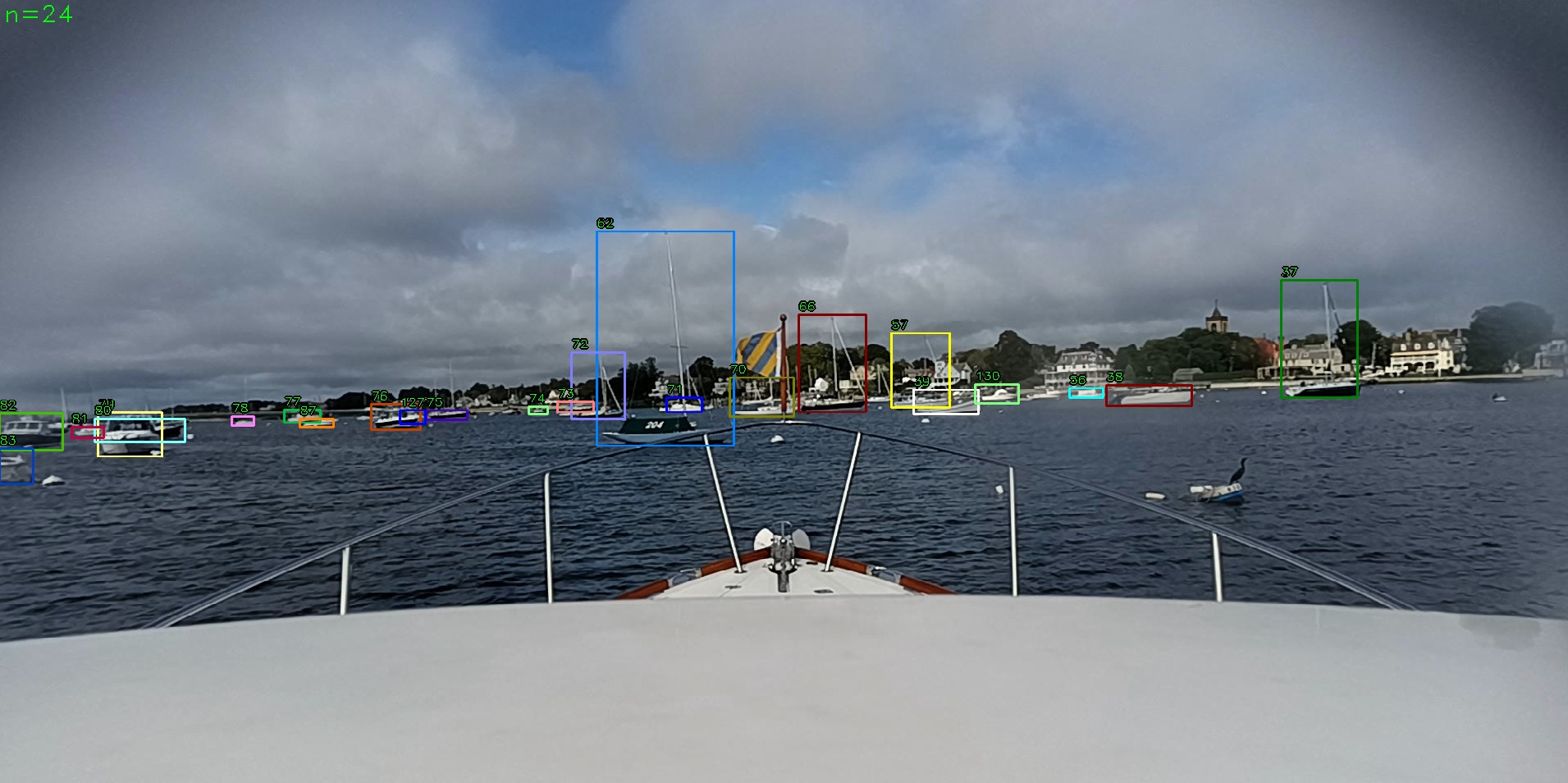}
   \caption{USV-based Multi-Object Tracking}
\label{fig:g1}
\end{subfigure}%
\hfill
\caption{Overview of MaCVi 2024 challenges.}
\label{fig:challenges_overview}
\end{figure}

In response to the growing interest and the need for specialized computer vision solutions tailored for the maritime domain, the first Workshop on Maritime Computer Vision (MaCVi)~\cite{Kiefer_2023_WACV} was organized in 2023.
The workshop highlighted research challenges, introduced standardized benchmarks brought together researchers and practitioners worldwide, fostering collaboration and driving innovation. The 2nd Workshop on Maritime Computer Vision (MaCVi2024), organized in conjunction with WACV2024, is a continuation of these efforts. MaCVi2024 introduces the following challenges as shown in Figure \ref{fig:challenges_overview}: UAV-based Multi-Object Tracking (MOT) w/ Re-identification, USV-based Obstacle Segmentation, USV-based Embedded Semantic Segmentation, USV-based Obstacle Detection, and USV-based MOT.





These competitions either are extensions from the last workshop (UAV-based MOT w/ ReID), employ new datasets (USV-based Obstacle Detection and Segmentation), or are new (USV-based MOT). Moreover, in response to the feedback and technological constraints highlighted in the previous MaCVi edition, this workshop features a new embedded sub-track within the Obstacle Segmentation challenge. 
The goal is to foster the development of 
lightweight, yet effective, computer vision algorithms suitable for deployment on devices with limited computational resources, commonly found on UAVs and USVs.

This report serves as a compass, guiding readers through the key findings, methodologies, and innovations presented during the workshop. The rest of the paper is organized as follows:
First, we provide an overview of the challenge protocol (Section 2), before we review the outcomes of the individual challenge tracks (Sections 3 and 4) with their underlying benchmarks and datasets.

\section{Challenge Participation Protocol}

The challenge tracks were announced and opened for participation on the 20th of September 2023. Participants were granted access to download the datasets, evaluation, and visualization toolkits from the workshop homepage\footnote{\url{https://macvi.org/workshop/macvi24}}. This allowed participants ample time to experiment with their methods on the provided data.

We opened the challenge server for submissions on the 20th of September 2023.
For all challenge tracks except the embedded one, the participants were required to uploaded their predictions for evaluation. For the embedded challenge, participants were required to export their models to ONNX for on-device benchmarking. The BoaTrack upload period started on the 18th of October.

To 
prevent overfitting, 
the participants were restricted to a single upload per day per challenge. 
After evaluation, the submissions appeared on the public leaderboard. Participants were offered the option to withdraw their submissions at any time. 
The challenge closed on the 3rd of November 2023, while new submissions after the 27th of October were hidden from others and only revealed after the conclusion of the challenges. 

To ensure a high standard of submissions, participants were informed that their submitted predictions would undergo further scrutiny. They were also mandated to provide comprehensive information about their methods. 
The pre-specified primary metrics for each challenge track determined the top-3 positions. Additionally, every participant was asked to provide information regarding the running time of their method, measured in frames per second wall clock time, their hardware specifications, and any extra datasets utilized (including those for pretraining).

In line with our commitment to advancing the state of maritime computer vision, 
top three ranked teams for all challenges were invited to submit a technical report detailing their methods and training configurations due by the 7th of November 2023. These insightful reports are attached in the appendix of this paper.
The results were presented at the MaCVi2024 workshop on the 7th of January 2024.

\subsection{Evaluation Server}

Following the announcement of the Maritime Computer Vision initiative, the evaluation server transitioned to a new domain at \texttt{macvi.org}. This move was not just a change of address; the webpage underwent a comprehensive restructuring to better accommodate the integration of new challenge tracks, ensuring scalability and ease of expansion for future challenges.

The current server is an evolution of the original web server used for the SeaDronesSee benchmark and has been readily accessible online several months prior to the challenge commencement. Alongside the main challenge tracks, it also supports additional tracks such as Boat-MNIST, UAV-based Object Detection v1/v2 and UAV-based Single-Object Tracking.

\section{UAV-based Object Tracking Challenge}
\label{sec:uavobjecttrackingchallenge}

The second iteration of the SeaDronesSee benchmark introduces an enhanced Multi-Object Tracking (MOT) track, now incorporating re-identification capabilities. This advancement builds on the initial focus of tracking objects in marine Search and Rescue (SaR) and surveillance scenarios; adding a critical layer of maintaining consistent subject identification. In this iteration, the challenge extends beyond tracking the movement and positions of people or boats over time to include re-identification, especially crucial for subjects that temporarily leave the field of view or become occluded.

The core challenges of tracking small, partly occluded subjects whose appearances change with movement and water-induced occlusion remain. Additionally, the complexities due to gimbal movement and altitude changes, causing rapid object motion within video frames, are addressed. The SeaDronesSee-MOT with re-identification track is designed to advance technologies in dynamic tracking and identification, essential for improving SaR and surveillance operations. The next section will explore the details and innovations of this updated challenge track.


\subsection{Dataset}

The updated SeaDronesSee-MOT dataset for the re-identification challenge encompasses enhancements, focusing primarily on the test set which now includes re-identification labels. This dataset comprises 21 train set clips, 17 in the validation set, and 19 in the test set, amounting to 54,105 frames.

As demonstrated in Table \ref{tab:unique_id_comparison}, the updated test set, enriched with re-identification labels, aims to track boats, swimmers, and floaters under a unified class setting, not distinguishing between these classes. This unified approach is pivotal for short-term tracking tasks, where objects that exit the scene are not tracked but are expected to be re-identified when they reappear. Naturally,the re-identification dataset often exhibits fewer unique object IDs than the standard MOT dataset. In instances where the re-identification dataset shows a greater number of unique IDs, this increase is attributed to the correction of previously mislabeled or unlabelled instances, enhancing the dataset's accuracy.

Additionally, each frame in this dataset is complemented by comprehensive metadata, including UAV altitude, gimbal angles, GPS coordinates, and more. This detailed metadata may aid in precise object tracking and identification.

\begin{table}
\centering
\caption{Comparison of Unique Object IDs in SeaDronesSee-MOT and -MOT with Re-identification Datasets on the SeaDronesSee-MOT test set. The table shows the number of unique object IDs for each video ID in both datasets (video IDs 7, 8 and 20 do not exist).}
\label{tab:unique_id_comparison}
\begin{tabular}{lcc}
\toprule
 Video ID &  Unique IDs &  Unique IDs (ReID) \\
\midrule
0 & 11 & 12 \\
1 & 15 & 15 \\
2 & 5 & 5 \\
3 & 10 & 10 \\
4 & 120 & 34 \\
5 & 11 & 11 \\
6 & 11 & 10 \\
9 & 14 & 11 \\
10 & 6 & 7 \\
11 & 3 & 3 \\
12 & 8 & 8 \\
13 & 6 & 6 \\
14 & 4 & 4 \\
15 & 10 & 4 \\
16 & 10 & 10 \\
17 & 5 & 5 \\
18 & 28 & 27 \\
19 & 1 & 1 \\
21 & 139 & 23 \\
\bottomrule
\end{tabular}
\end{table}

\subsection{Evaluation Protocol}

We evaluate the submissions by using the following metrics: HOTA, MOTA, IDF1, MOTP, MT, ML, FP, FN, Recall, Precision, ID Switches, Frag \cite{luiten2021hota,leal2015motchallenge}. The determining metric for winning is HOTA. In case of a tie, MOTA is the tiebreaker.

Furthermore, we require every participant to submit information on the computational runtime of their method measured in frames per second wall-clock time along their used hardware.

\subsection{Submissions, Analysis and Trends}

\begin{table*}[t]
\centering
\caption{Multi-Object Tracking with Re-identification submissions overview. For brevity, we denoted all=train and val set.}
\label{tab:mot_submissions_overview}
\vspace{-.2cm}
\begin{tabular}{lrrrrrr}
\toprule
Model name & Data & Detector & FPS & GPU  \\
\midrule
MG-MOT (UWIPL) (\ref{tr:seayoulater}) \cite{Yang_2024_WACV} & COCO, SeaDronesSee-MOT$^{\text{all}}$ & YOLOv8-x   & 13 & GV100 \\
Tracking by Detection (Fraunhofer IOSB) (\ref{tr:tbd}) & COCO, SeaDronesSee-MOT$^{\text{all}}$ &  Varifocal-Net   & 1 & V100  \\
ReIDTracker-Sea (Lenovo) (\ref{tr:reidtracker-sea}) & SeaDronesSee-MOT$^{\text{all}}$ & Swin-Transformer   & 3 & A100  \\
\bottomrule
\end{tabular}
\end{table*}

We received 49 submissions from 10 different teams. Additionally, we provided the same baseline as last time, i.e.~a Tracktor-based tracker using ECC with a Faster R-CNN ResNet-50 detector (\ref{tr:baseline}) without re-identification. 

40 submitted trackers outperformed the baseline. However, we are going to restrict our analysis on the best tracker of each of the best three teams.

See an overview of the submitted methods in Table \ref{tab:mot_submissions_overview}. Table \ref{tab:mot_results} shows the results of the best submissions of the best five teams. Again, the winner submissions followed the tracking-by-detection paradigm. Interestingly, winning submissions each employed a different object detector backbone: a one-stage YOLOv8-x, a two-stage Varifocal Net (ResNet-50), and a Transformer.

\begin{table*}[tb]
\centering
\caption{Multi-Object Tracking results on the SeaDronesSee-MOT test set. The submissions are ranked based on HOTA.}
\label{tab:mot_results}
\vspace{-.2cm}
\begin{tabular}{lrrrrrrrrrrrrr}
\toprule
Model name & HOTA & MOTA & IDF1 & MOTP & MT & ML & FP & FN & Re & Pr & IDs & Frag \\
\midrule
\color{gold(metallic)} MG-MOT & \color{gold(metallic)} 0.695 & 0.771 & 0.859 & 0.208 & 153 & 27 & 9531 & 12351 & 0.871 & 0.897 & 18 & 783 \\
\color{silver} TBD & \color{silver} 0.693 & 0.780 & 0.844 & 0.205 & 165 & 20 & 10643 & 10391 & 0.891 & 0.889 & 16 & 984 \\
\color{bronze} ReIDTracker-Sea & \color{bronze} 0.624 & 0.781 & 0.713 & 0.204 & 150 & 32 & 9595 & 11166 & 0.883 & 0.898 & 178 & 1343 \\
\bottomrule
\end{tabular}
\end{table*}

\begin{figure*}[tb]
\centering
   \includegraphics[width=\textwidth]{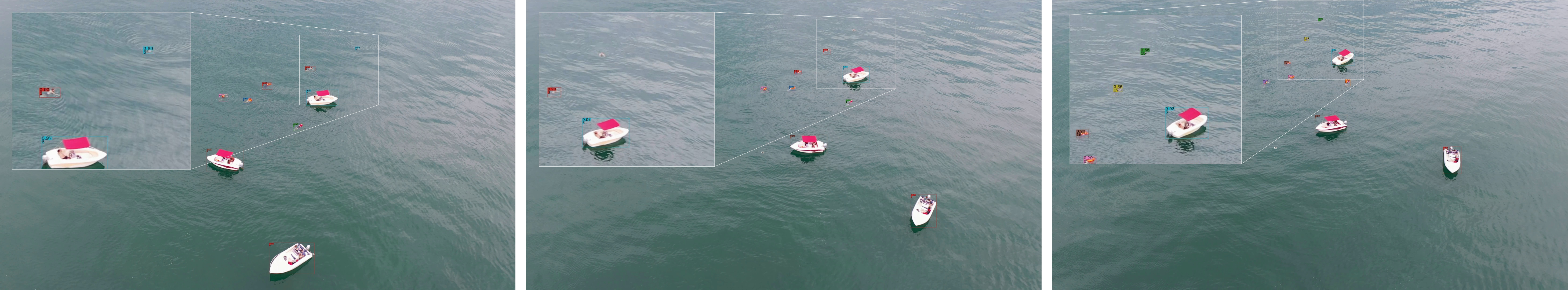}
      \includegraphics[width=\textwidth]{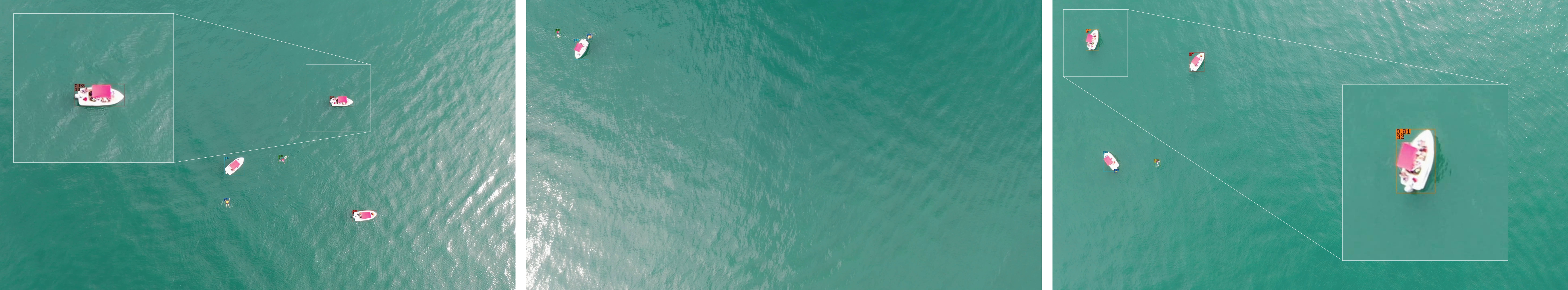}
\caption{Common fragmentation and reidentification errors (here \emph{ReIDTracker-Sea}). {\bf Top:} The detected swimmer in the first frame is lost in one of the next frames due to a heading angle change and is assigned a new ID when it is re-detected in a later frame. {\bf Bottom:} A boat is not re-detected re-entering the scene after having left it. This may be caused by the great similarity between the different boats. }
\label{figMOT:overviewtrackingerrors}
\end{figure*}

\emph{MG-MOT} (\ref{tr:seayoulater}) and \emph{TBD} (\ref{tr:tbd}) performed almost on-par in both metrics, HOTA and MOTA. While \emph{MG-MOT} has significantly fewer fragmentions, \emph{TBD} has the lowest number of identity switches (only 16). The third place, \emph{ReIDTracker-Sea} (\ref{tr:reidtracker-sea}) is on-par in terms of MOTA, but considerably worse in HOTA and IDF1, indicating its inferiority associating objects. In fact, its number of ID switches and fragmentions are considerably higher, while precision and recall values are at the top. This is illustrated in Figure \ref{figMOT:overviewtrackingerrors}, showing a common fragmention and re-identification error caused by a sudden movement of the camera. In this particular case, we suspect the missing motion model of \emph{ReIDTracker-Sea} to be the cause for this as both, \emph{MG-MOT} and \emph{TBD} don't do these errors.

While these errors where caused by camera movements in the \emph{short-term}, a common source of errors was caused by very similar looking boats in the \emph{long-term}. Figure \ref{figMOT:overviewtrackingerrors} shows how a boat is assigned an incorrect ID after it has left the scene and re-entered it. This is a special case of long-term tracking where it is particularly hard for any metadata-agnostic re-identification module to re-identify the instance correctly. The winning model \emph{MG-MOT} is the only submission that does not assign incorrect IDs in this specific example. This is likely due to its metadata-guided module that leverages metadata to obtain an understanding of the objects' topology.

\subsection{Discussion}

The UAV-based Multi-Object Tracking with Reidentification challenge highlighted challenges in differentiating similar objects from high altitudes during fast camera movements. The winning team, \emph{MG-MOT}, successfully incorporated onboard metadata to address these issues. \emph{Tracking by Detection}, in second place, also showcased strong performance with a well-tuned existing framework. The third-place team, \emph{ReIDTracker-Sea}, utilized a Transformer for effective detection, though the absence of a motion model may have affected their association accuracy.


\section{USV-based Perception Challenges}
\label{sec:usvchallenges}


The following USV-oriented subchallenges were organized: (i)
obstacle detection, (ii) obstacle segmentation, (iii) multi-object tracking, and (iv) the embedded segmentation challenge. The latter is the first of its kind,
dedicated to promoting development of perception methods capable of running on embedded hardware.
The challenges used two datasets: the LaRS benchmark~\cite{Zust2023LaRS} was used in segmentation and detection challenges, while the multi-object tracking challenge applied a dataset BoaTrack datast, captured by LOOKOUT \cite{lookout}. Both of these datasets feature scenes captured from the viewpoint of USVs and boast a large scene and obstacle variety (see Figure~\ref{fig:challenges_overview}).


\subsection{Datasets}

\subsubsection{LaRS}

\begin{figure}
\centering  
   \includegraphics[width=\linewidth]{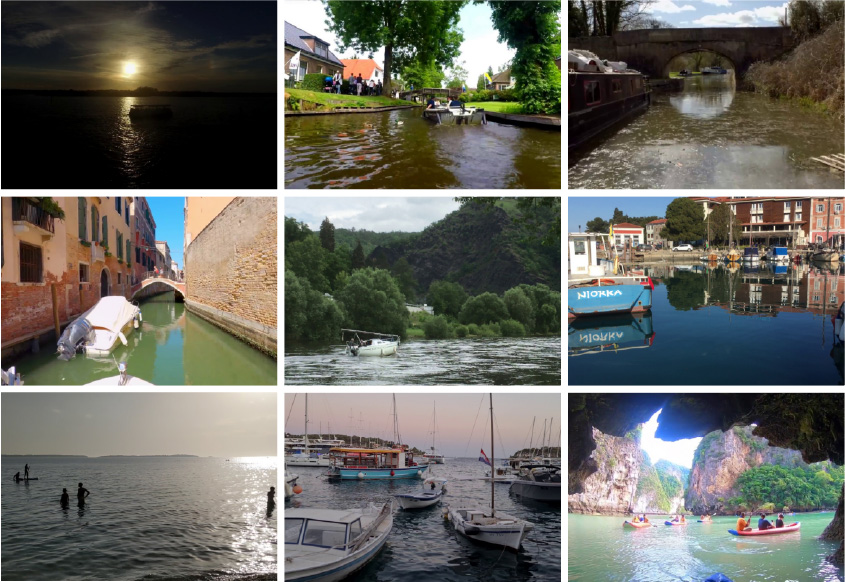}
\caption{Examples of scenes in the LaRS benchmark.}
\label{fig:lars-images}
\end{figure}

The obstacle segmentation, embedded obstacle segmentation and obstacle detection challenges, used the recently released LaRS benchmark~\cite{Zust2023LaRS}. It contains over 4000 challenging and visually diverse maritime and inland scenes (see Figure~\ref{fig:lars-images}) with panoptic obstacle annotations, featuring locations around the world. The panoptic annotations contain 3 stuff categories (sky, water and static obstacles) and 8 different dynamic obstacle instance categories (e.g. boats, buoys, swimmers). In addition, LaRS is annotated for semantic segmentation, where all static and dynamic obstacle categories are merged into a single obstacle segmentation mask. 

LaRS features challenging scenarios such as scenes with object reflections, sun glitter and bad visibility. To this end, scenes have been additionally labeled with several scene-level attributes to allow for detailed analysis. LaRS is split into train, validation and test sets. The annotations for the train and validation sets are publicly available, while the annotations of the test set are withheld, to ensure fair comparison of methods. Instead, an online evaluation server is hosted to evaluate user-submitted predictions. For more details on the data acquisition and annotation processes refer to \cite{Zust2023LaRS}.

\subsubsection{BoaTrack}

\begin{table}[tb]
\centering
\caption{BoaTrack statistics. The average lifespan indicates how long an ID lives across frames before it leaves the scene.}
\label{tab:combined_gt_stats}
\vspace{-.2cm}
\begin{tabular}{lccc}
\toprule
Video & \# Objects & Unique IDs & Avg. Lifespan\\
\midrule
Video 1  & 36,116 & 130 & 277.82 \\
Video 2  & 1,180 & 3 & 393.33 \\
Video 3  & 66,072 & 94 & 702.89 \\
\midrule
\textbf{Total} & \textbf{103,368} & \textbf{227} & \textbf{455.37} \\
\bottomrule
\end{tabular}
\end{table}

BoaTrack is a new multi-object tracking (MOT) dataset recorded from the viewpoint of USVs. The goal of BoaTrack is to advance computer vision algorithms in autonomous boating and boating assistance systems. It is aimed at detecting and tracking boats and other objects (such as buoys) in open water and dock scenarios. For the purposes of the MaCVi challenge, only boats are to be tracked. Participants needed to submit object locations and IDs for each frame.

 The focus is on short-term tracking, i.e., re-identification of objects that leave and re-enter the field of view is not required. 

The test set contains three videos clips, totaling 10,656 frames. Please note, that we did not provide any train or val videos. This means, participants needed to train a detector on other datasets, such as the aforementioned LaRS, but were free to use any other publicly available dataset for training. Please see an overview of the dataset in Table \ref{tab:combined_gt_stats}. There are 8,255 frames for the first video and 901 and 1,500 for the second and third, respectively.

\subsection{USV-based Obstacle Segmentation Challenge}
\label{sec:usvobstaclesegmentation}

The methods participating in the USV-based Obstacle Segmentation Challenge were required to predict the scene segmentation (into obstacles, water and sky) for a given input image.
The submitted methods have been evaluated on the recently released LaRS benchmark~\cite{Zust2023LaRS}. In addition to the publicly available training set, the authors were also allowed to use additional datasets (upon declaration) for training their methods.

\subsubsection{Evaluation Protocol}
\label{usv-seg:evaluation}

To evaluate segmentation predictions, we employ the LaRS~\cite{Zust2023LaRS} semantic segmentation evaluation protocol. Segmentation methods provide per-pixel labels of semantic components (water, sky and obstacles). However, traditional approaches for segmentation evaluation (\eg mIoU) do not consider the aspects of predictions that are relevant for USV navigation. Instead, the LaRS protocol evaluates the predicted segmentations with respect to the downstream tasks of navigation and obstacle avoidance and focuses on the detection of obstacles. 

\begin{table*}
\caption{Performance of the submitted segmentation methods and baselines (denoted in gray) on the LaRS test set. Performance is reported in terms of water-edge accuracy ($\mu$), precision (Pr), recall (Re), F1 score, segmentation mIoU and overall quality (Q = mIoU $\times$ F1).}
\label{tab:usvseg-results}
\centering
\begin{tabular}{cllcccccc}
\toprule
 Place &         Method & Institution &          \bf{Q} $\downarrow$ &   $\mu$ &         Pr &         Re &         F1 &       mIoU \\
\midrule
\mfirst{} &    SWIM$^2$ [\S\ref{usv-seg:swim2}] &  UniCa & \bm1{78.1} & \bm1{79.7} & \bm3{76.9} & \bm1{83.0} & \bm2{79.9} & \bm2{97.8} \\
\msecond{} & TransMari [\S\ref{usv-seg:transmari}]  &     HKUST & \bm2{77.8} & \bm2{79.6} & \bm2{78.5} & \bm2{82.0} & \bm1{80.2} &       97.1 \\
\mthird{} & Mari-Mask2Former [\S\ref{usv-seg:marimask2former}] &      DLMU & \bm3{75.7} &       78.4 & \bm1{79.7} &       75.1 & \bm3{77.3} & \bm2{97.8} \\
\midrule
4th &      Mask2Former &       HSU &       73.8 &       78.5 & \bm3{76.9} &       73.8 &       75.3 & \bm1{98.0} \\
-   & \color{gray} K-Net & \color{gray} MaCVi &       71.3 & \bm3{78.8} &       67.6 & \bm3{80.4} &       73.4 & \bm3{97.2} \\
-   & \color{gray} DeepLabv3 & \color{gray} MaCVi &       62.9 &       77.5 &       61.1 &       72.0 &       66.1 &       95.2 \\
5th &            eWaSR &  EAIC-UIT &       56.5 &       67.8 &       55.5 &       62.7 &       58.9 &       96.0 \\
6th &        OneFormer &    DLR-MI &       52.0 &       68.3 &       47.4 &       62.7 &       54.0 &       96.2 \\
7th &           PidNet &      XJTU &       50.7 &       74.7 &       47.0 &       62.8 &       53.7 &       94.3 \\
\bottomrule
\end{tabular}
\end{table*}

The detection of static obstacles (\eg shoreline) is measured by the water-edge accuracy ($\mu$), which evaluates the segmentation accuracy around the boundary between the water and static obstacles. On the other hand, the detection of dynamic obstacles (\eg boats, buoys, swimmers) is evaluated by counting true-positive (TP), false-positive (FP) and false-negative (FN) detections, summarize by the F1 score. A ground-truth obstacle is counted as a TP if the intersection with the predicted obstacle segmentation is sufficient, otherwise it is counted as a FN. FPs are counted as the number of segmentation blobs (after connected components) on areas annotated as water in ground truth. For further details please see \cite{Zust2023LaRS}.

The detection F1 score is a great indicator of the quality of the method predictions as all obstacles have equal importance in the final score, regardless of their size. However, in the trivial case of predicting everything as an obstacle, all ground-truth obstacles will be counted as TP and there will be only one FP blob per image (albeit very large), which leads to a very large F1 score. On the other hand, mIoU measures the overall segmentation quality on a per-pixel level, but does not reflect the detection of smaller obstacles very well. 
We thus combine the two measures into a single quality measure (Q = F1 $\cdot$ mIoU) and use it as the primary performance measure in this challenge.

\subsubsection{Submissions, Analysis and Trends}

\begin{figure*}[t]
\centering
   \includegraphics[width=\linewidth]{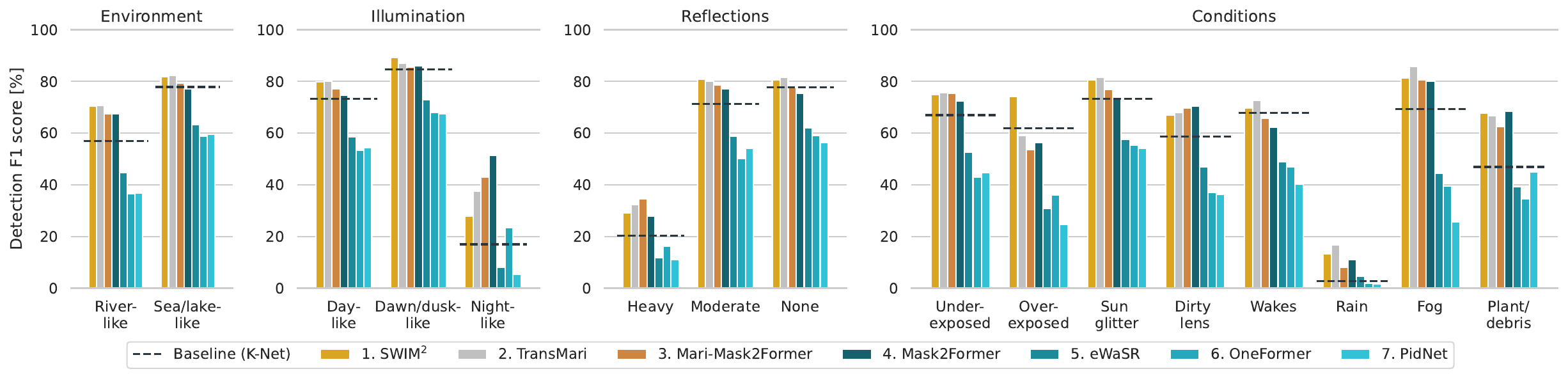}
\caption{Performance of top segmentation methods with respect to different scene attributes. The performance of the baseline method K-Net is marked as a dotted line for reference.}
\label{fig:usvseg-cat}
\end{figure*}

\begin{figure}[t]
\centering
   \includegraphics[width=\linewidth]{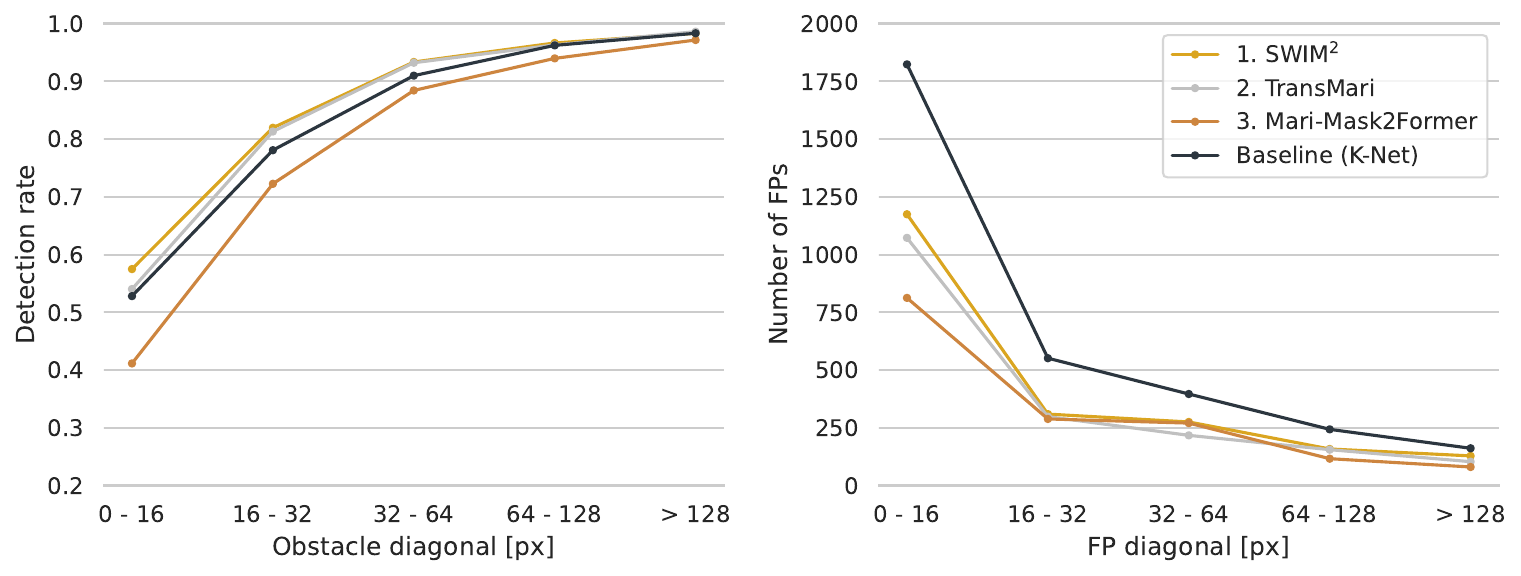}
\caption{Obstacle detection rate and number of FPs across different sizes of obstacles for top performing methods in comparison to the baseline.}
\label{fig:usvseg-sizes}
\end{figure}

\begin{figure*}
    \centering
        \includegraphics[width=\textwidth]{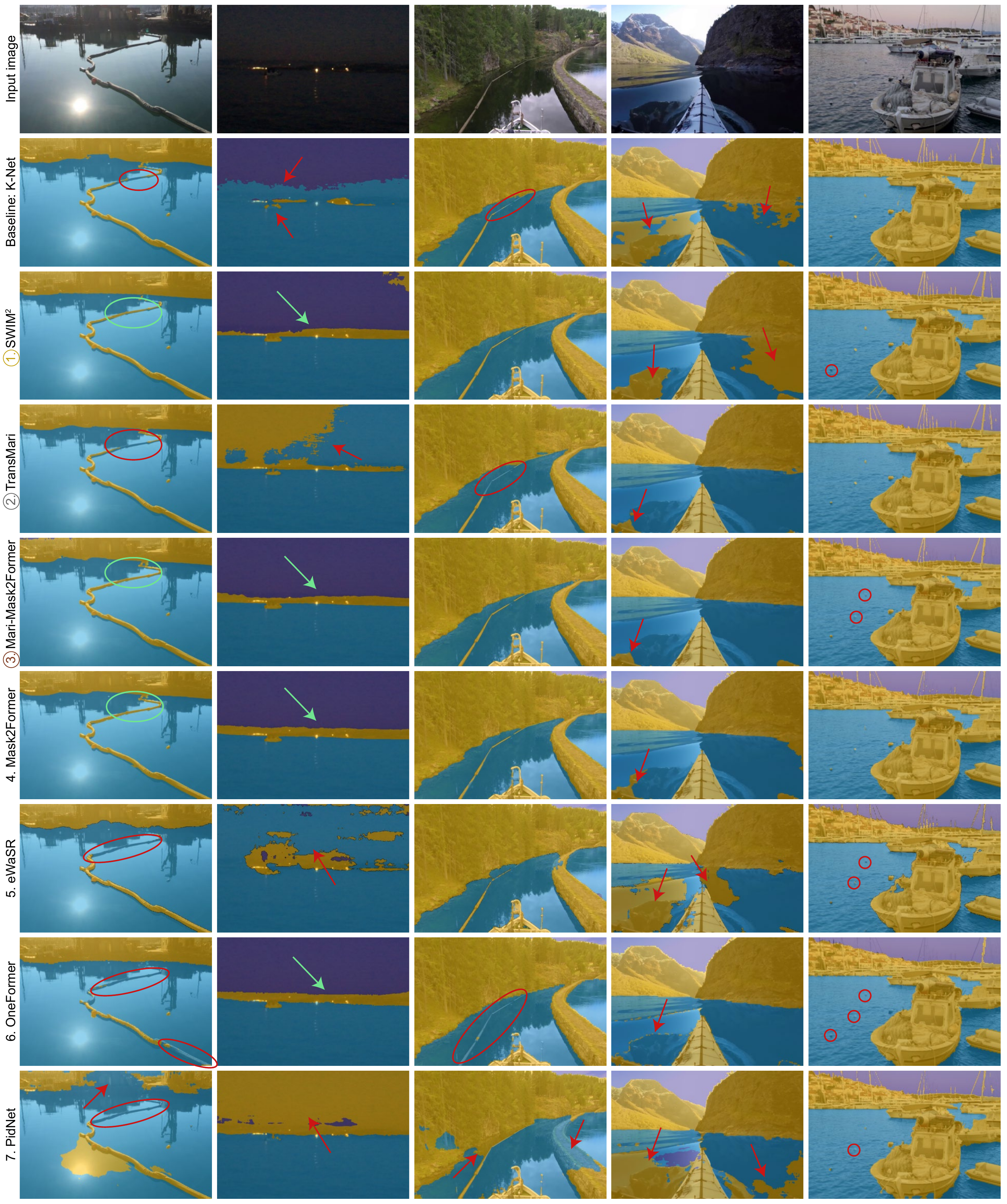}
    \caption{Qualitative comparison of methods for USV-based Obstacle Segmentation. Prominent errors and examples of good behaviour are highlighted.}
\label{fig:usvseg-qualitative}
\end{figure*}

The USV-based obstacle segmentation challenge  received 49 submissions from 7 different teams. 
As per the rules of the challenges, only the best-performing method from each team is considered in the following analysis. Results of the remaining submitted methods are available on the public leaderboards of the challenge on the MaCVi website~\cite{MaCViWebsite}.
Table~\ref{tab:usvseg-results} presents the overall standings of the teams participating in the challenge and the results of their best performing methods. In addition to the competing teams, we also analyze two baselines provided by the MaCVi 2024 committee, namely DeepLabv3 and K-Net, which was the previous state-of-the-art on the LaRS benchmark~\cite{Zust2023LaRS}. In this section we analyze all the contributed methods, with a special focus on the top three approaches. We refer to teams and their methods by their overall ranking in Table~\ref{tab:usvseg-results} with the notation \rankn{$n$}, where $n$ is the place of the approach. 
Short descriptions of the top three methods are available in the Appendix~\ref{sec:reports/usvseg}.

Overall, four teams, \rankn{1} UniCa, \rankn{2} HKUST, \rankn{3} DLMU and \rankn{4} HSU, improved over the previous state-of-the-art performance of K-Net~\cite{Zhang2021KNet} by quite a large margin. Namely, the top performing method SWIM$^2$ outperforms it by 6.8\% Q. Interestingly, all these methods are based on the popular Mask2Former architecture~\cite{cheng2021mask2former} with slight alterations. Specifically, SWIM$^2$ adapts the training strategy and 
insights from biomedical computer vision 
to improve the recognition of small objects, which are common in LaRS and Mari-Mask2Former utilizes the dice loss, to address the issue of class imbalance. The top-two methods, SWIM$^2$ and TransMari, achieved quite a similar performance, with SWIM$^2$ outperforming TransMari by only 0.3\% Q. TransMari performed slighly better in F1 detection score, but ranked lower in overall segmentation quality (mIoU).

Real-time semantic segmentation methods eWaSR~\cite{tervsek2023ewasr} and PidNet~\cite{xu2022pidnet} were also considered by two teams, but have been outperformed by computationally more intensive methods. \rankn{6} utilized the recent OneFormer~\cite{Jain2023OneFormer} architecture for universal semantic segmentation, but opted to use a lower processing resolution compared to other methods. None of analysed methods utilized additional data during training. In the following, we analyze the best-performing methods of each of the teams in more detail.

\textbf{Detection by scene attributes:} 
Figure~\ref{fig:usvseg-cat} 
shows the performance 
with respect to scene attributes, including environment type, illumination, amount of reflections and scene conditions. Overall, the top four methods outperform the baseline (K-Net~\cite{Zhang2021KNet}) in most categories, particularly in 
challenging scenarios such as heavy reflections, foggy scenes and plants or debris presence in water. SWIM$^2$ demonstrates remarkable stability across different scenarios and is the only method that consistently outperforms the baseline in all categories.

\textbf{Performance by obstacle size:} 
Figure~\ref{fig:usvseg-sizes} 
compares the detection performance of the
top three methods with the K-Net~\cite{Zhang2021KNet} baseline across different obstacle sizes. The largest differences between methods are revealed on small obstacles. 
Specifically, we observe that all three methods 
significantly reduce the number of small false positives, while preserving high detection rates across the board. In particular, both SWIM$^2$ and TransMari achieve higher detection rates than the baseline across all obstacle sizes, while consistently reducing the number of false positives. Mari-Mask2Former, on the other hand, mainly outperforms the baseline through increased precision, as seen in a drastic reduction of FPs (especially small ones), but detects fewer obstacles overall.

\textbf{Qualitative results:} Examples of predicted scene segmentations are presented in Figure~\ref{fig:usvseg-qualitative}. In contrast to the baseline we observe remarkable robustness on night scenes (column 2) and increased segmentation accuracy on thin structures (columns 1 and 3) for methods SWIM$^2$, Mari-Mask2Former and Mask2Former. OneFormer, which employs a similar architecture, also performs well in night scenes, but is unable to accurately segment thin structures (columns 1 and 3) or smaller objects (column 5) due to operating at a lower resolution. Furthermore, all methods still have some trouble generalizing to varied reflections such as the example shown in column 4.

\subsubsection{Discussion and Challenge Winners}

The overall winners of the USV-based Obstacle Segmentation challenge are:
\begin{description}
    \item[1\textsuperscript{st} place:] University of Cagliari with SWIM$^2$,
    \item[2\textsuperscript{nd} place:] Hong Kong University of Science and Technology (HKUST) with TransMari, and
    \item[3\textsuperscript{rd} place:] Dalian Maritime University (DLMU) with Mari-Mask2Former
\end{description}
All three methods significantly outperformed the previous state-of-the-art on the LaRS benchmark and demonstrate remarkable robustness to various challenging scenarios such as night scenes, small objects and thin structures. The 1\textsuperscript{st} and 2\textsuperscript{nd} placing SWIM$^2$ and TransMari methods set a new state-of-the-art for semantic segmentation on LaRS. The detection of tiny objects and robustness to novel water reflections remain open issues and we expect that the performance will continue to improve as new methods are developed to tackle these problems.


\subsection{USV-based Embedded Obstacle Segmentation}
\label{sec:usvembeddedobstaclesegmentation}

Recent state-of-the-art obstacle detection methods typically utilize computationally expensive and power-hungry hardware, which makes them unsuitable for small-sized energy-constrained USVs~\cite{tervsek2023ewasr}. To bridge this gap, the focus should be put on optimizing existing methods and developing new techniques. Thus, this challenge is an extension of the USV-based Obstacle Segmentation challenge described in Section~\ref{sec:usvobstaclesegmentation}, with the main goal of developing an obstacle segmentation method suitable for deployment on an embedded device. The methods are run, benchmarked, and evaluated on a real-world device -- an upcoming next-gen device from Luxonis based on Robotic Vision Core 4 (RVC4).

\subsubsection{Evaluation Protocol}

Since the methods need to be suitable for deployment on an embedded device, we introduce additional constraints that need to be considered during development and submission.  The following criteria and rules need to be respected:
\begin{itemize}
    \item \emph{Static graph} -- the neural network must have a defined static graph and exported to ONNX so that it can be successfully compiled for the target platform.
    \item \emph{Supported operations} -- target platform supports a limited set of supported operations, which we provide before the beginning of the challenge.
    \item \emph{Standardized inputs} -- since models are evaluated on the device, we request that models expect input images of fixed size ($768 \times 384$) and must be normalized with ImageNet~\cite{deng2009imagenet} mean and standard deviation. We only allow methods with a single image input.
    \item \emph{Throughput} -- achieved throughput on the embedded device must be $\ge 30$ FPS.
\end{itemize}

Submitted methods are first quantized to the INT8 format on the validation set of LaRS~\cite{Zust2023LaRS} and compiled to a binary that can be executed on the target device. Before inference, all images are resized, centered, and padded to $768 \times 384$ shape with preserved aspect ratio, and the outputs are resized to the original image resolution. Then, the same evaluation protocol as in \ref{usv-seg:evaluation} is used to derive the final scores. In addition to the LaRS metrics, 
the average throughput achieved on the embedded device is also reported.

\subsubsection{Submissions, Analysis and Trends}

35 submissions from 3 different teams, including 8 baseline models from the MaCVi2024 committee were evaluated. We show the baseline models and the best submission from each team in Table~\ref{tab:usveseg-overview}. As stated in the challenge rules, only models that are faster than the predetermined threshold of $30$ frames per second were considered for the final rankings.

\begin{table*}[h]
\centering
\caption{Overview of the submissions for the USV-based Embedded Obstacle Segmentation challenge. When determining the final placement of the teams, methods that achieve the required $\ge 30$ FPS on the target device are considered (denoted in teal). For evaluation comparison, we include the winning method from the non-embedded challenge from Section \ref{sec:usvobstaclesegmentation} at the top, but do not consider it as part of the challenge. The best results from considered methods are denoted in bold.}
\label{tab:usveseg-overview}
\begin{tabular}{lcrrccccccc}
\toprule
Place & Institution & Method & Section & FPS & $\textbf{Q}\downarrow$ & $\mu$ & Pr & Re & F1 & mIoU \\
\midrule
           & \color{gray} \textit{UniCa} & \color{gray} \textit{SWIM$^2$}              & \ref{usv-seg:swim2}             & /                            &  78.1 & 79.7	& 76.9	& 83.0	& 79.9	& 97.8 \\ \midrule
           & \color{gray} UL      & \color{gray} SegFormer (MiT-B2)  	   & \ref{usv-eseg:baselines}        & \color{red} 15.0                         &  55.8 & 69.0	& 50.9	& 67.4	& 58.0	& 96.3 \\
\mfirst{}  & EAIC-UIT             & eWaSR-RN50          	   & \ref{usv-eseg:ewasr}            & \color{teal} \textbf{103.4} &  \textbf{55.3} & \textbf{68.5}	& 48.5	& \textbf{70.5}	& \textbf{57.4}	& \textbf{96.2} \\
           & \color{gray} UL      & \color{gray} DeepLabv3+ (RN-101)& \ref{usv-eseg:baselines}        & \color{red} 16.6                        &  54.9 & 68.7	& 56.2	& 58.6	& 57.4	& 95.7\\
           & \color{gray} UL      & \color{gray} FCN (RN-101)      & \ref{usv-eseg:baselines}        & \color{red} 16.8                         &  51.5 & 68.6	& 53.5	& 54.1	& 53.8	& 95.6\\

\msecond{} & DLMU                 & Mari-MobileSegNet  		   & \ref{usv-eseg:marimobilesegnet} & \textcolor{teal}{60.3}  &  51.3 & 69.1	& \textbf{55.7}	& 52.2	& 53.9	& 95.3\\
           & \color{gray} UL      & \color{gray} CN (RN-50)        & \ref{usv-eseg:baselines}        & \color{red} 19.7                         &  50.9 & 68.0	& 53.1	& 55.0	& 54.0	& 94.2\\
           & \color{gray} UL      & \color{gray} STDC2             & \ref{usv-eseg:baselines}        & \textcolor{teal}{38.3}                        &  47.5 & 67.6	& 50.5	& 49.2	& 49.9	& 95.3\\
           & \color{gray} UL      & \color{gray} STDC1             & \ref{usv-eseg:baselines}        & \textcolor{teal}{45.8}  &  45.1 & 66.8	& 47.2	& 48.5	& 47.9	& 94.3\\
           & \color{gray} UL      & \color{gray} BiSeNetv2         & \ref{usv-eseg:baselines}        & \textcolor{teal}{44.6}  &  42.8 & 64.7	& 43.2	& 49.0	& 46.0	& 93.2\\
           & \color{gray} UL      & \color{gray} BiSeNetv1 (RN-50)  & \ref{usv-eseg:baselines}        & \color{red} 28.7                         &  42.6 & 63.8	& 38.2	& 56.6	& 45.6	& 93.4\\
\bottomrule
\end{tabular}
\end{table*}

\begin{figure}[h]
\centering
   \includegraphics[width=\linewidth]{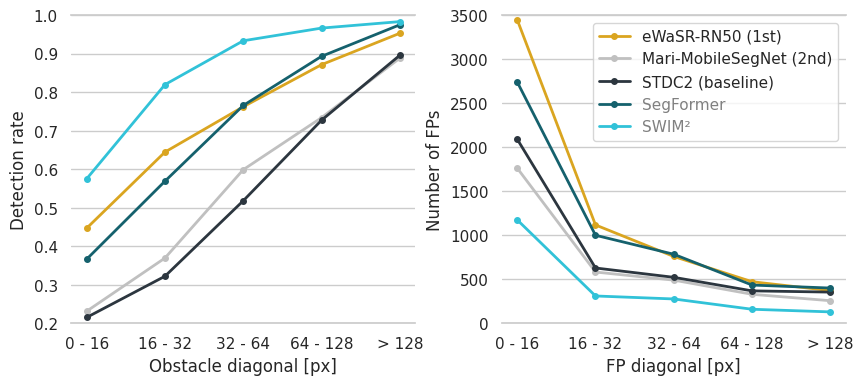}
\caption{Detection rate and number of false positive predictions by diagonal size. We show the best three methods within FPS constraints and the best method regardless of the FPS from \ref{sec:usvembeddedobstaclesegmentation}, and the best method from \ref{sec:usvobstaclesegmentation} for comparison.}
\label{fig:usveseg-sizes}
\end{figure}

\begin{figure*}[ht]
\centering
   \includegraphics[width=0.85\linewidth]{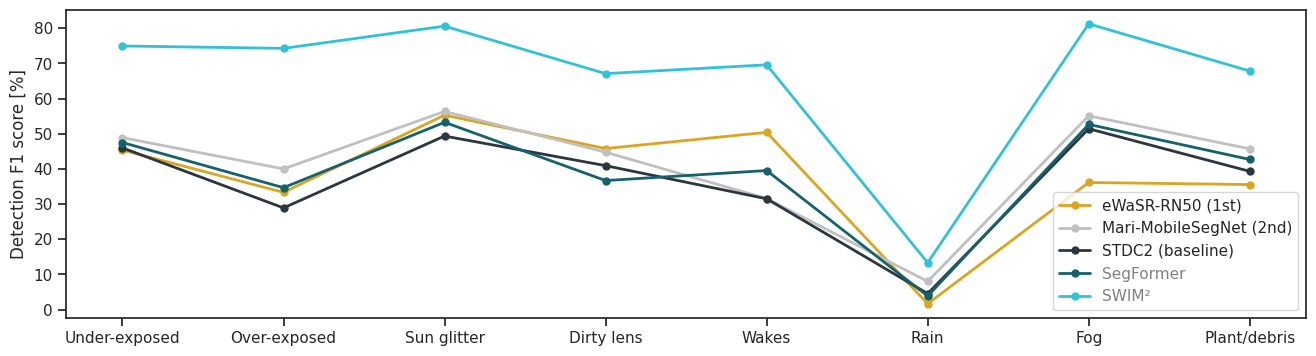}
\caption{Detection F1 score under different conditions. We show the best three methods within FPS constraints and the best method regardless of the FPS from \ref{sec:usvembeddedobstaclesegmentation}, and the best method from \ref{sec:usvobstaclesegmentation} for comparison.}
\label{fig:usveseg-conditions}
\end{figure*}

\begin{figure*}[t]
    \centering
        \includegraphics[width=\textwidth]{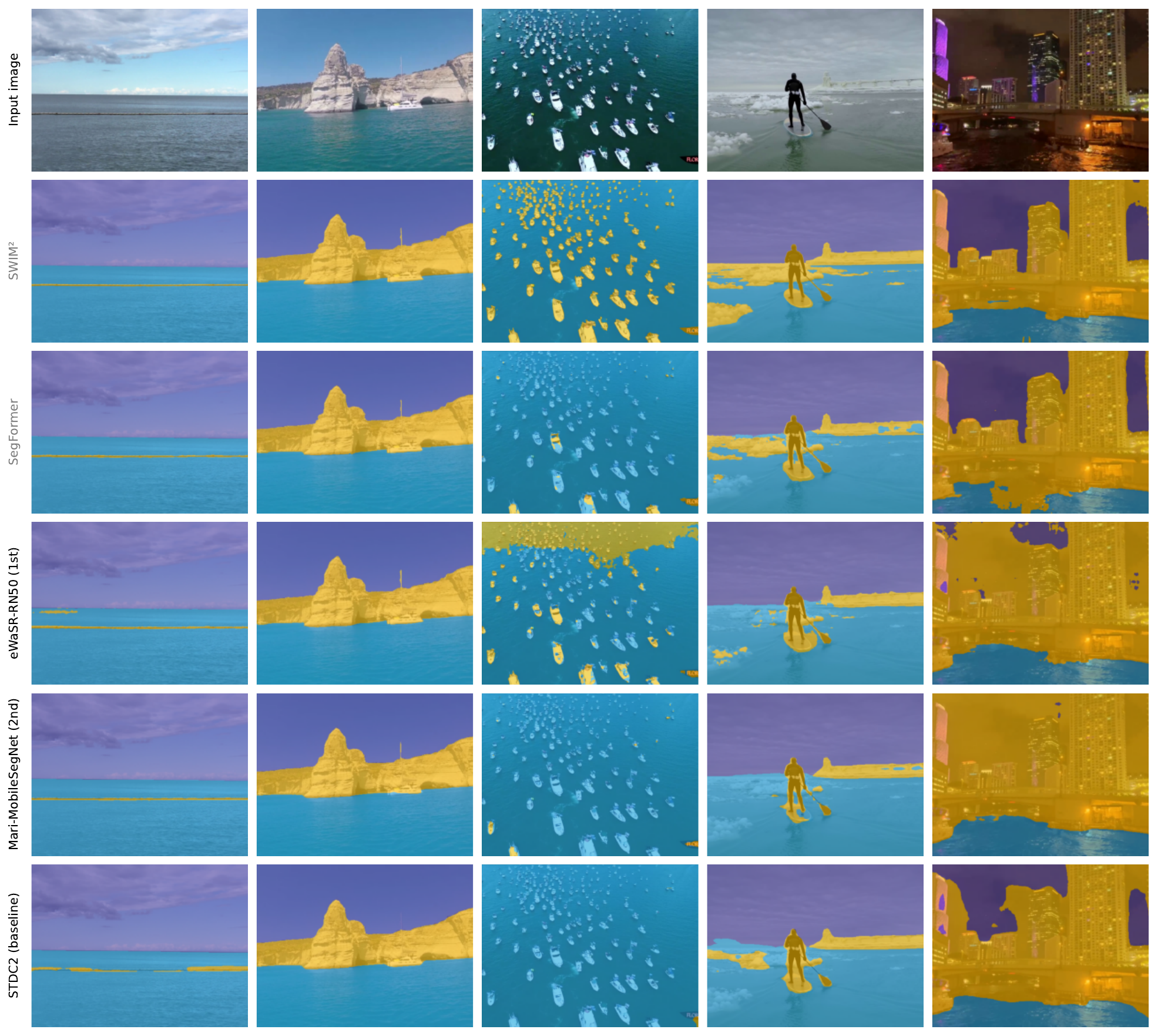}
    \caption{Qualitative comparison of methods for USV-based embedded obstacle segmentation.}
\label{fig:usveseg-qualitative}
\end{figure*}

The submitted methods explored various architecture changes and training regimes. eWaSR-RN50 (Section~\ref{usv-eseg:ewasr}) is based on eWaSR~\cite{tervsek2023ewasr}. 
The authors fine-tuned various hyper-parameters, resulting in an increase of the batch size from 4 to 6, replacing the ResNet-18~\cite{he2016deep} backbone with a heavier ResNet-50, and increasing the training duration. Furthermore, resizing all input images to a fixed size of $768 \times 384$ boosts performance, compared to center cropping during training.

Mari-MobileSegNet (Section~\ref{usv-eseg:baselines}) utilizes the MobileNetV2 backbone~\cite{sandler2018mobilenetv2}, a typical architecture for mobile devices. The decoder employs the ASPP decoding head and is based on DeepLabV3~\cite{chen2017rethinking} architecture. FCNHead~\cite{long2015fully} is used as an auxiliary head during training.

Compared to the baselines (Section \ref{usv-eseg:baselines}) from the MaCVi2024 committee that satisfy the $30$ FPS requirement, both, eWaSR-RN50 and Mari-MobileSegNet achieve higher throughput and Q score. While eWaSR-RN50 dominates in most metrics, Mari-MobileSegNet achieves higher precision (+$7.2\%$) at the cost of recall ($-18.3\%$). Compared to the best baseline method SegFormer, which does not satisfy the FPS requirements, eWaSR-RN50 achieves only $0.5$ lower Q score, while being almost $7 \times$ faster on an embedded device.

We investigate how SegFormer, eWaSR-RN50, Mari-MobileSegNet, and the best baseline within FPS limits (STDC2) compare to the best method from the non-embedded segmentation challenge (SWIM$^2$) from Section~\ref{sec:usvobstaclesegmentation}. We observe that SWIM$^2$ is $+22.8$ Q better than eWaSR-RN50. This can be attributed to the smaller input shape that the embedded methods use at inference time, leading to poor performance on very small obstacles (see Figure~\ref{fig:usveseg-sizes}). SWIM$^2$ archives almost $3 \times$ better detection rate on obstacles with diagonal length between $0$ and $16$ pixels compared to Mari-MobileSegNet. This difference is less noticeable for SegFormer and Mari-MobileSegNet, but both produce more small-sized false positives.

In Figure~\ref{fig:usveseg-conditions}, we show that the detection F1 score under different conditions follows a similar trend for the $5$ analyzed models. While the detection performance gap is noticeable between the non-embedded method and the other embedded methods, all struggle in the presence of rain, but are most robust to the presence of sun glitter and fog. From the embedded methods, both SegFormer and eWaSR-RN50 perform significantly better in the presence of wakes, however the latter struggles the most in the fog compared to the others.

We show qualitative differences for the $5$ methods in Figure \ref{fig:usveseg-qualitative}. We can see that eWaSR-RN50 is the best embedded method for detecting small obstacles such as distant boats, but at the same time, it produces more false positive detections. SegFormer is better on ice and sky segmentation. Mari-MobileSegNet and STDC2 struggle the most with small obstacles. As expected, the non-embedded SWIM$^2$ performs the best on all examples.

\subsubsection{Discussion and Challenge Winners}

The overall winners of the USV-based Embedded Obstacle Segmentation are:
\begin{description}
    \item[1\textsuperscript{st} place:] University of Information Technology, VNU-HCM (EAIC-UIT) with eWaSR-RN50, and 
    \item[2\textsuperscript{nd} place:] Dalian Maritime University (DLMU) with Mari-MobileSegNet.
\end{description}

The analysis indicates that the detection performance of 
the embedded-ready methods falls short from the non-embedded methods. 
This primarily stems from the fixed lower-resolution input, which makes it hard to achieve a good detection rate on small obstacles. However, 
the detection rate gap is significantly reduced for large obstacles.


\subsection{USV-based Obstacle Detection Challenge}
\label{sec:usvobstacledetectionchallenge}

\begin{table*}[t]
\centering
\caption{Overview of the submissions for the USV Obstacle Detection challenge, with results. Ranking of the method on the leaderboard as well as the final placement of the teams are indicated. We also include the self-reported inference speeds, used hardware, and whether any other datasets were used in the training.}
\label{tab:usvdet-overview}
\begin{tabular}{clllllll}
\toprule
Place & \# & Author & Model Name & FPS & Hardware & Add. data & F1 \\
\midrule
\mfirst{} & 1 & FER-LABUST, Croatia & YOLOv8x, pretrained [\S \ref{usv-det:YOLOv8ExtraLarge}] & 40 & GTX 1080 TI & Yes & 51.50 \\
\msecond{} & 3 & Fraunhofer IOSB & Co-DETR [\S \ref{usv-det:Co-DETR}] & 4 & A100 & Yes & 43.56 \\
\mthird{} & 6 & DLR MI-SIT & ScatYOLOv8CBAM+SAHI [\S \ref{usv-det:ScatYOLOv8CBAM}] & 4 & A100 &  & 39.79 \\
4th & 14 & User \#382 & yolo\_test2 & 0.1 & 1080 &  & 35.87 \\
5th & 15 & DLR MI-SIT  & ModelSahi & 1.5 & A6000 &  & 34.21 \\
6th & 18 & mcprl & codetr\_0.2 & 1 & 1080 &  & 22.13 \\
- & 20 & Baseline UL & YOLO v7 baseline & 10 & 1080 Ti &  & 18.16 \\
\bottomrule
\end{tabular}
\end{table*}

This year's obstacle detection relies on the the newly released LaRS Dataset~\cite{Zust2023LaRS}, which is primarily a segmentation dataset. However, with this challenge we recognize that for early practical applications of USV collision avoidance and path planning, robust dynamic obstacle detection is beneficial, even if lacking per-pixel accuracy.

Thus in the Obstacle Detection track, the task was to develop an obstacle detection method capable of identifying obstacles in the input image and representing their location using rectangular bounding boxes. This was in accordance with the LaRS Dataset nomenclature, which identifies 8 classes of dynamic obstacles (boat, buoy, other, row boat, swimmer, animal, paddle board, float) and three classes of "stuff", referring to pixels in the image that do not correspond to any of the dynamic obstacle classes. For the purpose of this challenge, all eight LaRS dynamic obstacle classes were treated as a single obstacle class and given equal importance.

\subsubsection{Evaluation protocol}

The detection performance was evaluated using the standard Intersection over Union (IoU) metric for bounding box detections. A detection was classified as a true positive (TP) if it achieved an IoU score of at least 0.3 with the ground truth. Conversely, detections were considered false positives (FP) if they did not meet this threshold, unless 75\% or more of the pixels in the submitted bounding box overlapped with the image area labeled as "static obstacle" as defined in the LaRS Dataset nomenclature. This approach effectively meant that false positive detections on land were not a concern. The final score was calculated as an average F1 score, derived from TP and FN metrics.

Additionally, every participant was required to submit information on the speed of their method, measured in frames per second (FPS). This also included an indication of the hardware used for benchmarking the speed. Lastly, participants were asked to disclose which datasets, including those used for pretraining, were utilized during the training process.

\subsubsection{Submissions, Analysis and Trends}
We received many submissions, the challenge was far more popular than MaCVi 2023. The top performing submissions are reported in Table~\ref{tab:usvdet-overview}. Sorted by $F1$ metric, the top of the list is dominated by three teams, FER LABUST, Fraunhofer IOSB and DLR MI-SIT, who were invited to submit descriptions of their approach, which are provided in sections \ref{usv-det:YOLOv8ExtraLarge}, \ref{usv-det:Co-DETR} and \ref{usv-det:ScatYOLOv8CBAM}, respectively.

\subsubsection{Discussion and Challenge Winners}

The winners of the USV object detection challenge are as follows:
\begin{description}
    \item[1\textsuperscript{st} place:]{FER LABUST with YOLOv8x,}
    \item[2\textsuperscript{nd} place:]{Fraunhofer IOSB with PRBNet,}
    \item[3\textsuperscript{rd} place:]{DLR MI-SIT with ScatYOLOv8CBAM+SAHI}
\end{description}

 An important difference from last year's challenge is that a YOLO-based method took the crown, helped by additional pre-training on several datasets.

\subsection{USV-based Multi-Object Tracking}

Analogously to the SeaDronesSee-MOT challenge track, we evaluate the submissions on HOTA, MOTA, IDF1, MOTP, MT, ML, FP, FN, Recall, Precision, ID Switches, Frag \cite{luiten2021hota,leal2015motchallenge}. The determining metric for winning is HOTA, MOTA is the tiebreaker. We also required every participant to submit information on the computational runtime of their method measured in frames per second wall-clock time along their used hardware.

\subsubsection{Submissions, Analysis and Trends}

\begin{table*}[t]
\centering
\caption{BoaTrack submissions overview. }
\label{taboatrack_submissions_overview}
\vspace{-.2cm}
\begin{tabular}{lrrrrrr}
\toprule
Model name & Data & Detector & FPS & GPU  \\
\midrule
Detector Ensemble (UWIPL) (\ref{tr:UWIPLboatrack}) & COCO, LaRS & YOLOv8-x   & 6.6 & V100 \\
ReIDTracker-Sea (Lenovo) (\ref{tr:reid_boatrack}) & COCO, LaRS &   Swin-Transformer  & 1 & V100  \\
DLR-BoaTrack (DLR) (\ref{tr:dlr_team}) & LaRS & ScatYOLOv8+CBAM   & 3 & A100  \\
\bottomrule
\end{tabular}
\end{table*}

\begin{table*}[tb]
\centering
\caption{Multi-Object Tracking results on the BoaTrack test set. The submissions are ranked based on HOTA.}
\label{tab:motboatrackresults}
\vspace{-.2cm}
\begin{tabular}{lrrrrrrrrrrrrr}
\toprule
Model name & HOTA & MOTA & IDF1 & MOTP & MT & ML & FP & FN & Re & Pr & IDs & Frag \\
\midrule
\color{gold(metallic)} D. Ens. (\ref{tr:UWIPLboatrack} & \color{gold(metallic)} 0.215 & 0.094 & 0.247 & 0.232 & 9 & 162 & 13231 & 80383 & 0.222 & 0.635 & 83 & 407 \\
\color{silver} ReIDT. (\ref{tr:reid_boatrack}) & \color{silver} 0.214 & 0.105 & 0.232 & 0.214 & 17 & 131 & 14476 & 76651 & 0.258 & 0.649 & 1404 & 4185 \\
\color{bronze} DLR (\ref{tr:dlr_team}) & \color{bronze} 0.193 & 0.057 & 0.202 & -1.000 & 7 & 173 & 12275 & 85058 & 0.177 & 0.599 & 183 & 541 \\
\bottomrule
\end{tabular}
\end{table*}

Despite the later release and start of the challenge, we received 23 submissions from 8 different teams. Interestingly, the performance in terms of the main metrics, HOTA, MOTA, and IDF1, is significantly worse compared to the UAV-based MOT with ReID challenge. We suspect this to be mainly due to the train-test-set domain gap as participants had to train on a dataset with likely significantly different characteristics. Also, the high number of objects per frame poses a difficulty (compare to Figure \ref{tab:combined_gt_stats}). Also see Figure \ref{fig:challenges_overview} for a sample scene, showing that there are both, large foreground and small background objects.

Interestingly, ReIDTracker-Sea is in the top three of both competitions, UAV-based MOT with Reidentification and USV-based MOT, indicating its robustness to different domains.

\section{Conclusion}

In this summary paper, we analyzed the challenges of the 2\textsuperscript{nd} Workshop on Maritime Computer Vision. Specifically, MaCVi 2024 hosted one aerial and four different surface domain maritime challenges, some of which featured new exciting benchmarks for maritime vision. 

The UAV-based Multi-Object Tracking with Reidentification challenge showed that there are still challenging conditions when there are fast camera movements or similar looking object that are hardly distinguishable from high altitudes, making reidentification difficult without considering the topology of the scene. Using the onboard's metadata, this can be incorporated and this year's winner, \emph{MG-MOT}, did so. The second place \emph{Tracking by Detection} performed almost equally well with a well-known tracking framework, indicating that clean tuning of existing methods is still very promising as well. The third place, \emph{ReIDTracker-Sea} leverages a Transformer that has a good detection capability, but their method is missing a motion model, which likely has a negative impact on association accuracy.


The USV-based challenges featured two new datasets, LaRS and BoaTrack, each with unique challenges. Winners of the \emph{USV Obstacle Segmentation Challenge} were (1\textsuperscript{st}) SWIM$^2$ from University of Cagliari, (2\textsuperscript{nd}) TransMari from Hong Kong University of Science and Technology, and (3\textsuperscript{rd}) Mari-Mask2Former from Dalian Maritime University. All methods significantly outperformed the baseline, dramatically improving on issues such as night scenes and thin structures, with SWIM$^2$ becoming the new state-of-the-art for semantic segmentation on LaRS.

MaCVi 2024 also hosted an embedded segmentation challenge for the first time, with the goal of developing and assessing methods for obstacle segmentation suitable for use on real world embedded hardware. Winners of the \emph{USV-based Embedded Obstacle Segmentation Challenge} were (1\textsuperscript{st}) eWaSR-RN50 University of Information Technology, and (2\textsuperscript{nd}) Mari-MobileSegNet from Dalian Maritime University. Both methods achieve a high Q score compared to other embedded baseline models while running faster than $60$ FPS on an embedded device, while the 1\textsuperscript{st} method boasts superior detection of small obstacles. However, we still observe a large performance gap on all methods compared to state-of-the-art methods for obstacle segmentation and more effort needs to be invested into developing robust embedded-ready segmentation methods.

USV-based Object Detection was hosted again in a simplified form. It used the new LaRS dataset in an object detection scenario. In contrast to last year, a YOLO-based method took the first place, helped by additional pre-training.


The accuracy for the USV-based Multi-Object Tracking challenge is still very low compared to other MOT benchmarks. We believe this to be credited to the crowded nature of harbor scenes as indicated by the low precision and recall values. The winner, \emph{Detection Ensembe} employed an ensemble model for higher performance. The second place, \emph{ReIDTracker-Sea} performs well in this and the UAV-based competition, while the third place $\emph{DLR-BoaTrack}$ leverages an improved YOLOv8 model.

In summary, the outcomes of the MaCVi 2024 challenges underscore the advancing maturity of the maritime computer vision field. The increasing resilience of methods to prevalent issues like reflection reflects significant progress. However, real-world hardware limitations remain a notable constraint, revealing performance gap of embedded methods. Developing efficient methods for addressing this challenge is becoming an important frontier for enabling real world applications of autonomous maritime technology.

\vspace{.2cm}
\noindent\textbf{Acknowledgments.}
This work was supported by
Conservation, Protection and Use joint call,
Slovenian Research Agency (ARRS) project J2-2506 and programs P2-0214 and P2-0095,
Sentient Vision Systems for sponsoring prizes for the UAV-based Multi-Object Tracking challenge, and LOOKOUT for providing testing videos for the USV-based Multi-Object Tracking challenge.


\appendix
\addcontentsline{toc}{section}{Submitted Methods}
\section*{Appendix - Submitted Methods}


\section{UAV-based Tracking}

\subsection{MG-MOT: Metadata-Guided Long-Term Re-Identification for UAV-Based
Multi-Object Tracking}
\label{tr:seayoulater}
\emph{Cheng-Yen Yang, Hsiang-Wei Huang, Zhongyu Jiang, Heng-Cheng Kuo,
Jie Mei, Jenq-Neng Hwang}\\
\emph{cycyang@uw.edu}\\
\\
{\bf 1. Method:} Metadata-Guided MOT
Our proposed method Metadata-Guided MOT (MG-MOT) takes the metadata provided along with the data to
do the long-term re-identification to recover and resulted in
a much improved IDF1 of the MOT result. Most details can
be found in our paper \cite{Yang_2024_WACV} and please feel free to use any of
the figures in our work in the challenge report!\\
{\bf 2. Dataset:}
We used the entire training and validation dataset of
SeaDroneSee-MOT \cite{varga2022seadronessee} to train our multi-class detector
with the pretrained weight on COCO. Metadata of the
training, validation, and testing are also used as a source to
select a set of reasonable thresholds for our algorithm.\\
{\bf 3. Environment:}
Our codebase is derived from ultralytics \cite{yolov8github}. We trained
our detector using 8 Nvidia V100 GPUs and the inference
step including predicting the bounding boxes and online
tracking is done on a single Nvidia GV100 GPU and i7-
10700K CPU @ 3.80GHz. The rough estimate of the latency is 13 FPS and given the tracks of the dataset are not
much, the post-processing time is neglectable compared to
the remaining latency.\\
{\bf 4. Additional Findings \& Thoughts:}
We tried using the altitude to fine-tune our detector but
turned out not helping too much.
We also found some problems in the dataset (e.g., annotations and metadata): (1) There exists a certain amount of
ID labeling error in training and validation despite the tracks
not leaving the image at all. (2) There exists some synchronization problems and some data errors in the metadata of
the drone.
It is also quite surprising that there are still great num-
bers of FP and FN on the testing data, it will be nice if the
organizer can visualize some of our and other participants
results to show the error. \emph{Acknowledged by the authors of the dataset. Hope this is resolved with this paper or the actual workshop event.}

\subsection{Tracking-by-Detection (TBD)}
\label{tr:tbd}
\emph{Daniel Stadler, Lars Sommer}\\
\emph{\{daniel.stadler, lars.sommer\}@iosb.fraunhofer.de
}\\
\\
{\bf Tracking-by-Detection (TBD)}\\ Eponym of our approach is the \textit{tracking-by-detection} (TBD) paradigm which separates the multi-object tracking task into two sub-tasks: detection and tracking. Details of the applied methods for detection, tracking, and additional post-processing are given in the following.

\textbf{Detector:} To generate our detections, we used VarifocalNet \cite{vfnet} and ResNet-50 \cite{resnet} as backbone architecture. 
For initialization, we used weights pre-trained on MS COCO \cite{coco}. 
SGD was used as optimizer with an initial learning rate of 0.02, a momentum of 0.9, and a weight decay of 0.0001. 
The model was trained for 12 epochs. 
We employed the SeaDronesSee Object Detection v2 train and validation set as training data. 
For images with dimensions less than 3840x2160 pixels, we used multiple scales (1920x1080, 2376x1296, 2688x1512, and 3360x1890). Otherwise, we set the input scale to 3360x1890 pixels. For inference, we applied multi-scale testing (2688x1512, 3360x1890, and 4032x2268).
We considered all five classes during training and inference.
The implementation provided by MMDetection~\cite{mmdetection} – an open source object detection toolbox based on PyTorch – was used to train our detector. 
We used 2 Tesla V100 GPUs (CPU: Intel Xeon E5-2698 v4 @ 2.20GHz). 
The inference speed of the detector was about 1 FPS.

\textbf{Tracker:} We followed a two-stage IoU-based association scheme in which first, high-confident detections are matched to all tracks, and second, low-confident detections are matched to the remaining unassigned tracks similar as in \cite{bytetrack}.
To prevent the start of FP tracks, detection boxes with aspect ratio larger than 5 were removed.
Furthermore, an occlusion-aware initialization method was applied that filters detections with high overlaps to already tracked targets.
We used the NSA Kalman filter \cite{giaotracker} as motion model.
Moreover, ORB features \cite{orb} were extracted and matched in consecutive frames to estimate transformation matrices that were used for camera motion compensation.

\textbf{Post-processing}: The resulting tracks were linearly interpolated and short tracks with less than 13 detections were removed.
To merge tracks of the same object that couldn't be tracked at once because the object temporarily left the scene, we developed a hierarchical clustering based on appearance features extracted with the model from \cite{botsort}.
For each track, a mean feature vector was computed with the appearance features extracted from the tracks' detections.
The cosine similarity between all tracks was calculated and two tracks were merged if their similarity score was above a threshold, where most similar tracks were merged first.
Also, temporal constraints were enforced to prevent infeasible merges. 
The overall pipeline inference speed was less than 1 FPS.

\subsection{ReIDTracker Sea: BoaTrack \& SeaDronesSee}
\label{tr:reidtracker-sea}
\emph{Kaer Huang, Aiguo Zheng, Weitu Chong, Kanokphan Lertniphonphan, Jun Xie, Feng Chen, Jian Li, Zhepeng Wang}\\
\emph{\{huangke1, zhengag, klertniphonp, xiejun, chenfeng13, lijian30, wangzpb\}@lenovo.com, wtzhong22@m.fudan.edu.cn}
\\
Our maritime Multi-Object Tracking (MOT) solution for UAVs and USVs simplifies traditional complex MOT systems by using unsupervised learning with self-supervision on ImageNet and high-quality detectors. This approach, less reliant on costly video annotations, achieved top performances in major UAV and USV-based MOT competitions.

{\bf In our introduction}, we explore achieving state-of-the-art (SOTA) results in Multi-Object Tracking (MOT) using only high-performance detection and appearance models. Our approach employs CBNetV2 Swin-B \cite{liang2022cbnet} for detection and MoCo-v2 \cite{he2020momentum} for a self-supervised appearance model. We omit traditional motion information like the Kalman filter and IoU mapping and utilize ByteTrack \cite{bytetrack} to associate detection boxes of varying scores. For UAV datasets with low object overlap, we adjusted the NMS threshold accordingly.

\begin{figure}
  \centering
  \includegraphics[width=0.45 \textwidth]{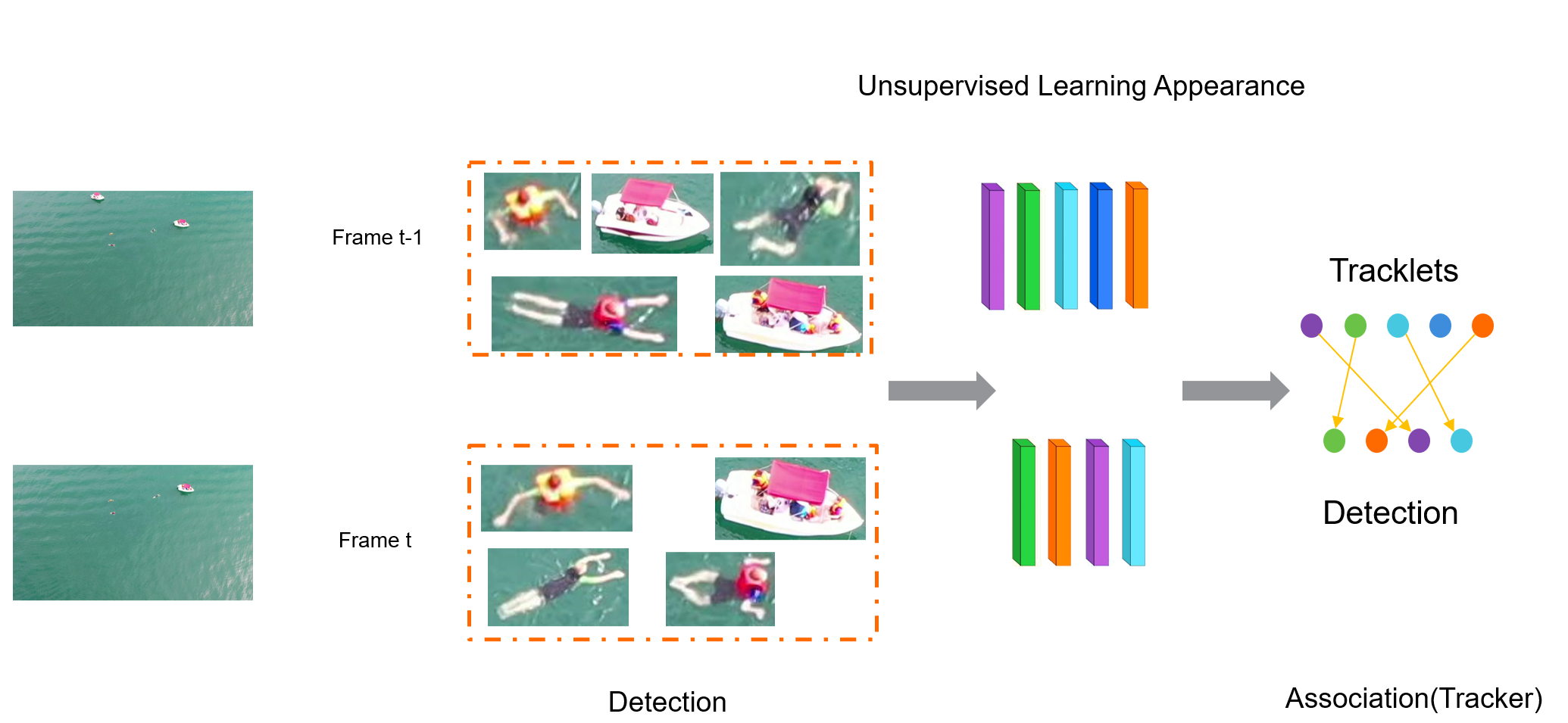}
  \caption{Overall architecture of ReIDTracker Sea.
  }
  \label{fig:mot_reid_3rd}
\end{figure}

{\bf Our MOT framework} includes Detection, Appearance Model, and Data Association (see Fig. \ref{fig:mot_reid_3rd}). The architecture comprises detection (providing high-quality instance boxes), an appearance model (offering quality embedding features), and a data association tracker that stabilizes trajectory outputs. The detection strategy uses a Swin-based transformer backbone \cite{liu2021swin} with CBNetV2 architecture, integrating multiple backbone features. The CBNetV2 integrates high
and low-level features of multiple backbones which connected in parallel. The Feature Pyramid Network (FPN) \cite{lin2017feature}
neck and Hybrid Task Cascade (HTC) \cite{chen2019hybrid} detector are attached and trained in each backbone as a main branch and an assistant branch. Only the main branch is used in the inference process.
We use more weight to bound box regression than classification in Loss Function for a more compact detection box which will benefit appearance model performance.

\begin{figure}
  \centering
  \includegraphics[width=0.45 \textwidth]{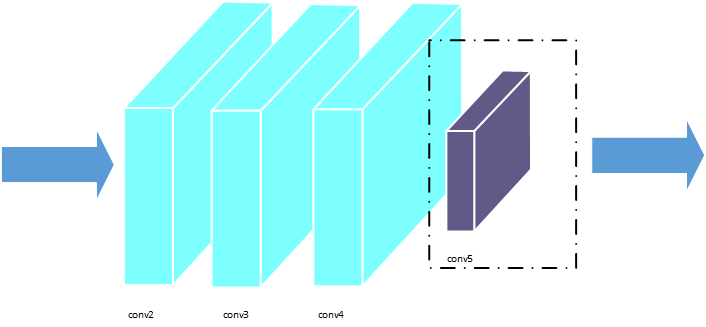}
  \caption{Appearance model of ReIDTracker Sea.
  }
  \label{fig:mot_reid_3rd_appearance}
\end{figure}

{\bf We use unsupervised appearance models} to address the high cost of video trajectory
annotation. Our base appearance model for this framework
is MoCo-v2 with ResNet50. The model extracts feature representations from detected boxes. MoCo-v2 model training by imagenet 1K dataset and then finetuning on MOT dataset. We also compare with model training by other contrastive learning methods (SimCLR, SimCLRv2 \cite{chen2020simple}, MoCo-v2, etc). We also make a comparison between supervised learning and self-supervised learning - we draw the conclusion that MoCo-v2 has better generalization capacity in the maritime dataset. Because the last convolution module of ResNet50 is more related to classification type, not the general features we want, we finally removed the network in the final integration
(Fig. \ref{fig:mot_reid_3rd_appearance}).

We adopt the Bytetrack concept, a simple but
strong method for {\bf matching object id across frames}. The
detected boxes in each frame are grouped based on their detection score into the high score and low score. Firstly, the method finds the association between the high score box and the tracklet. Then, the rest of the high score and low score
boxes are used to find the association from the remained
tracklet. The association method can be different in each
association step.
Our method uses only the appearance feature to associate
both high and low score boxes with tracklet. In addition, we
add a weighted score to tracklet to keep the tracklet representation from the higher detection score since the detection
score tends to get lower when the occluded part gets bigger.
The tracklet features are weighted by the detection score
and combined within $\tau$ frames to maintain the object representation during occlusion. The weighted feature $\hat{e}_j$ combined tracklet feature $e_j$ which is weighted by the detection score $s_j$ from the previous $\tau$ frames.
\[\hat{e}_j=\frac{\sum_{t=1}^\tau e_j^t \times s_j^t}{\sum_{t=1}^\tau s_j^t}\]
$\hat{e}_j$ is further used for finding the matched box in the data association. We apply the same association method
with \cite{wang2021different}. A ReId similarity matrix between tracklet and detection box is computed and used to find matching pairs by
the Hungarian algorithm \cite{kuhn1955hungarian}.

The Swin-B backbone was initiated by a
model pre-trained on ImageNet-22K. CBNetV2
was trained on the SeaDronesSee-MOT
train and val dataset. We applied multi-scale augmentation to scale the shortest side of images to between
640 and 1280 pixels and applied random flip augmentation during training. Adam optimizer was set with an initial
learning rate of 1e-6 and weight decay of 0.05. We trained
the model on 4 A100 GPUs with 1 image per GPU for 10
epochs. During inference, we resize an image to 2880x1920
to better detect the small objects. For the detection task, we
use a combination of classification Cross-Entropy loss and
the generalized IoU regression loss \cite{rezatofighi2019generalized}. Loss weights $\lambda_1$ and $\lambda_2$ are set to 1.0 and 10.0 by default, which drives the model output more compact box
\[\mathcal{L} = \lambda_1 \mathcal{L}_{cls}+\lambda_2 \mathcal{L}_{box}\]
The backbone of the appearance
model is pre-trained on ImageNet-1K. Then, we fine-tuned
the backbone by using MoCo-v2 on the SeaDronesSeeMOT dataset. The training dataset contains cropped object images according to bounding box labels from MOT dataset. The optimizer is SGD with a
weight decay of 1e-4, a momentum factor of 0.9, and an initial learning rate of 0.12. We trained the model on 4 A100
GPUs with 256 images per GPU.
Our method is generally similar to ByteTrack, but we used ReID to match high and low detection
boxes. We set the high detection score threshold to 0.84
and the low detection score threshold to 0.3.
for both challenges (SeaDroneSee-MOT and BoaTrack),
We use the sample ReID module which training on ImageNet 1K.
{\bf SeaDronesSee-MOT with Reid:} We just
train the detector using the SeaDronesSee-MOT train and
val set. but we grouped ”swimmer” and ”swimmer
with life jacket” and ”life jacket” as one class.
{\bf BoaTrack:} we just train the detector using LaRS train set
on ”boat” labels.

\subsection{Tracktor Baseline}
\label{tr:baseline}
\noindent
\emph{MaCVi Organizers}\\

The baseline is the last as in the last workshop iteration \cite{Kiefer_2023_WACV}. We provided a Tracktor-based tracker
using Camera Motion Compensation with a Faster R-CNN
ResNet-50 detector. We used the tracking implementations
from mmdetection \cite{mmdetection} with default hyperparameters.

\section{USV-based Obstacle Segmentation}
\label{sec:reports/usvseg}

\subsection{\protect\mfirst{} SWIM$^2$}
\label{usv-seg:swim2}
\noindent
\emph{Luca Zedda, Andrea Loddo, Cecilia Di Ruberto}\\
\texttt{\small{\{luca.zedda, andrea.loddo, cecilia.dir\}@unica.it,
}}\\
\emph{Department of Mathematics and Computer Science, University of Cagliari}\\
\textbf{Contributions:} Conceptualization: LZ, AL, CDR; Implementation: LZ; Experiments and; Analysis: LZ; Supervision: AL, CDR\\

\noindent
The submissions related to this report are known as Snarciv3, Snarciv2, and Snarci, with Snarciv3 being the 1$^{st}$ place winner of the competition. All architectures share the same backbone and model structure but vary in size and number of parameters. This technical report pertains to the winning solution: Snarciv3.

Our method is named SWIM$^2$, which stands for SWIn-based Maritime Mask2former. Our approach is based on the Mask2Former architecture~\cite{cheng2021mask2former}. We adapted the training strategy to use images of size 768x1536 to improve the recognition of small objects such as small boats and to better segment and recognize challenging objects found in real-life scenarios, such as those in the challenge dataset (e.g., algae and weeds growing on the borders of canals). We employed the Mask2Former base size model, which uses Swin as its backbone. For this downstream task, we chose to fine-tune the model that was pretrained on the Cityscapes dataset with a size of $512\times1024$. The original configuration file can be accessed at the following link\footnote{\url{https://github.com/open-mmlab/mmsegmentation/blob/main/configs/mask2former/mask2former_swin-b-in22k-384x384-pre_8xb2-90k_cityscapes-512x1024.py}}.

The preprocessing and augmentation pipeline includes the following steps: random resizing, cropping, flipping, and photometric distortion. These steps enhance data diversity and prepare it for robust model training. More details are available at the following link\footnote{\url{https://github.com/open-mmlab/mmsegmentation/blob/main/configs/mask2former/mask2former_r50_8xb2-90k_cityscapes-512x1024.py}}.

Additionally, we trained the model on a single Nvidia RTX 3060 with 12GB VRAM GPU, which led us to limit our batch size to 1. We trained our model for 90,000 iterations using a PolyLR scheduler and saved the best model based on the mIOU metric every 2000 iterations. The training took approximately 2 days, and the inference time averaged 3.1 images per second.

Our team primarily works with high-resolution images, specifically blood-smear images. These images share common issues with maritime images, such as the low pixel count of small parasitized cells (e.g., malaria-infected RBC) and their placement within low-entropy backgrounds. In addition, we also encounter variations in acquisition equipment, lighting conditions, factors such as object deformation, blurred boundaries, and color differences. These additional issues can make image analysis and processing even more challenging.

\subsection{\protect\msecond{} TransMari}
\label{usv-seg:transmari}
\noindent
\emph{Tuan-Anh Vu, Hai Nguyen-Truong, Tan-Sang Ha, Quan-Dung Pham, Sai-Kit Yeung}\\
\texttt{\small{\{tavu, thnguyenab, tsha, qdpham\}@connect.ust.hk, saikit@ust.hk,
}}\\
\emph{The Hong Kong University of Science and Technology }\\
\textbf{Contributions:} Conceptualization: TAVu, NTHai; Implementation: TAVu, TSHa; Experiments and Analysis: TAVu, QDPham; Supervision: SKYeung\\

\noindent
The proposed maritime obstacle segmentation is based on the transformer-based architecture, namely Mask2Former~\cite{cheng2021mask2former}. Mask2Former utilizes the identical meta-architecture as MaskFormer, except it substitutes the standard decoder with the new Transformer decoder with masked attention operator. It functions to extract localized features by restricting cross-attention to the foreground region of the predicted mask for each query rather than focusing on the entire feature map. A multi-scale approach is suggested as an effective means of managing small objects by leveraging high-resolution characteristics. Finally, they integrate optimization enhancements that augment the model's efficacy while preventing the need for additional computation. Our implementation based on MMSegmentation. The training configuration is adopted from the setting for ADE20K dataset~\cite{zhou2019semantic} with pretrained Swin-L backbone on ImageNet22k dataset~\cite{ILSVRC15}. The image resolution is set to $512\times512$, and the training schedule is set to “schedule 80k”. The optimizer is the AdamW optimizer, and the learning rate is 0.0001 with a momentum of 0.05. The data augmentation includes random crop and resize, random flipping, photometric distortion, noise transform, and normalization. We only use the LaRS dataset~\cite{Zust2023LaRS} for training, and no additional images are used. The training device is NVIDIA RTX 3090 with 24G memory. The inference speed is about 5.2 frames per second under the original image resolution of the dataset. The best submission of this method with test time augmentation achieved an average score of 77.8\% in the Q metric.

\subsection{\protect\mthird{} Mari-Mask2Former}
\label{usv-seg:marimask2former}
\noindent
\emph{Yuan Feng}\\
\texttt{\small{fengyuan@dlmu.edu.cn,
}}\\
\emph{Dalian Maritime University}\\

\noindent
Our method is named Mari-Mask2former and utilizes the relevant configuration files provided by MMSegmentation. The utilized model for USV obstacle segmentation is based on the transformer-based neural architecture Mask2former, which has shown outstanding performance in panoramic segmentation tasks. Inspired by this, we adapted the model in MMSegmentation by configuring it specifically for obstacle segmentation. The training configuration file is named \texttt{mask2former\_swin-l-in22k-384×384-pre\_8xb2\--160k\_ade20k-640×640} and it employs the Swin Transformer-Large as the backbone network. \\
\textbf{Backbone:} Swin-L pretrained on ImageNet-22k dataset with a resolution of 384×384 \\
\textbf{Decode head:} Mask2Former Head

\subsubsection*{Training}

\noindent
\textbf{Batchsize:} During the training process, the batch size used was 2. \\
\textbf{Optimizer:} Adamw, learning rate: 0.0001, eps: 1e-8, decay: 0.05, betas: (0.9 $\sim$ 0.999) \\
\textbf{Schedule:} We adopted a warm-up strategy and the "poly" learning rate scheduler during training. The training process lasted for a total of 160,000 iterations, starting from iteration 0 and ending at iteration 160,000. \\
\textbf{Augmentations:} including random flipping, random cropping, random brightness adjustment, and random saturation adjustment, among others. \\
\textbf{Loss functions:} We employed the following three loss functions: loss\_cls, loss\_dice, and loss\_mask. For loss\_cls, we used CrossEntropyLoss with a class weight of [1.0, 1.0, 1.0, 0.1], a loss weight of 2.0, a mean reduction and no sigmoid activation. For loss\_dice, we used DiceLoss with an activated True, an epsilon value of 1.0, a loss weight of 5.0, a naive\_dice calculation, a mean reduction, and sigmoid activation. Finally, for loss\_mask, we used CrossEntropyLoss with a loss weight of 5.0, a mean reduction, and sigmoid activation. \\
\textbf{Datasets used:} Backbone pretraining - ImageNet22k, Model pretraining - ADE20K 640×640, Finetuning - LaRs train \\
\textbf{Hardware:} The training device is NVIDIA RTX 3090 with 24G memory \\
\textbf{Inference Speed:} We tested LaRS's validation set using the code provided by MMSegmentation and obtained an average inference speed of 8.29 images per second.

\subsubsection*{Observations}

During training, we set up validation every 2400 iterations and output the Intersection over Union (IoU) and accuracy for each class. Upon analyzing all the validation results, we observed that the class of obstacles exhibited slow growth in both of these metrics. We hypothesized that this may be due to an imbalance in the distribution of pixels across obstacles, sky, and water in the images. Obstacles occupy fewer pixels, with the majority of them being located in the foreground of the images.

Inspired by this finding, we considered employing Dice loss to address the issue of sample imbalance. Dice loss effectively leverages foreground information. Additionally, we can also explore assigning higher weights to the obstacle class when using the cross-entropy loss function, which would allow the network to focus more on difficult-to-classify samples. However, it is important to be mindful that convergence issues, such as slow convergence or difficulty in convergence, may arise when using multiple loss functions simultaneously.

When using the Mask2former method, we observed that the timestamp in the testing images was recognized as an obstacle. We also tried using Deelplabv3+ with an R101 backbone network and found the same phenomenon. However, we did not observe this issue when using the Segformer method. Intuitively, algorithms that classify timestamps as obstacles may not perform well. Surprisingly, the predictions from Mask2former were more accurate than Segformer and Deelplabv3+. Although we have not conducted an extensive investigation into this matter, we find this observation to be quite interesting.

\subsection{Baselines}
\label{usv-seg:wasr-baseline}
\noindent
\emph{MaCVi Organizers}\\

\noindent
We provided two baselines for the challenge, DeepLabv3~\cite{chen2018deeplab} with the ResNet-101~\cite{he2016deep} backbone and K-Net~\cite{Zhang2021KNet} with the Swin-T~\cite{liu2021swin} backbone. We use the model implementations in the MMSegmentation framework~\cite{mmseg2020}. The methods were trained for 80k iterations on 2$\times$ NVIDIA V100 GPUs with a batch size of 8. We employ random resizing, flipping, photo-metric distortions for image augmentation. We apply random cropping with 1024$\times$512 crop size during training. During inference all input images are scaled to the 2048$\times$1024 resolution.

\section{USV-based Embedded Obstacle Segmentation}
\label{usv-eseg:submissions}

\subsection{\protect\mfirst{} eWaSR-RN50}
\label{usv-eseg:ewasr}
\noindent
\emph{Nguyen Thanh Thien}\\
\texttt{\small{thiennt@uit.edu.vn
}}\\
\emph{University of Information Technology}\\

\noindent
This report describes our solution for the USV Embedded Obstacle Segmentation Challenge. We achieve first place with $57.4$ F1 and $96.2$ mIoU, which results in a Q (Quality) score of $55.3$ on the leaderboard.

\noindent\textbf{Method:} Our submission is based on eWaSR~\cite{tervsek2023ewasr}. Its architecture includes three parts: backbone, decoder, and segmentation head. The backbone is the only component that we change: the original backbone ResNet-18~\cite{he2016deep} is replaced by larger version ResNet-50. With the new backbone, features needed for the decoder must be changed.
Specially, features from layers 10, 19, 31 and 49, which are selected because their output features have the same size as output from layers 6, 10, 14, 18 of ResNet-18 backbone, are used.

\noindent\textbf{Training:} eWaSR~\cite{tervsek2023ewasr} official repository is used for training and exporting the model. We trained the model for 100 epochs with batch size 6. Albumentations~\cite{albumentations} is applied to resize training images to 384x768 (see more explaination in Observations section below). Other settings are kept as default.

\noindent\textbf{Datasets:} We use LaRS dataset~\cite{Zust2023LaRS} for training, no other dataset is utilized.

\noindent\textbf{Hardware}: A P100 16GB GPU is used for training.

\noindent\textbf{Observations:}
\begin{itemize}
    \setlength{\itemsep}{1pt}
    \setlength{\parskip}{0pt}
    \setlength{\parsep}{0pt}
    \item Longer training improves the model’s performance. The original eWaSR model (with ResNet-18 backbone)
achieves 41.6 Q with only 25 epochs of training. The score increases to 46.6 and 47.1 Q with 50 and 100 epochs of
training, respectively.
    \item Larger backbone enhances the result. With 100 epochs of training, model with ResNet-50 backbone obtains the best result with 54.1 Q, while model with ResNet-34 backbone is slightly worser with 52.8 Q. Although using large backbone reduces the inference speed (from 117.3 to 103.8 FPS), in this case, it is very worthwhile because the result is still good compared to the required threshold (30 FPS).
    \item Augmentations may give an undetermined impact. For most of submissions, our team resizes the training images by combining two Albumentations~\cite{albumentations} transforms: LongestMaxSize and CenterCrop. Using this combination, we get the result 54.1 Q with ResNet-50 backbone. Later, we use normal Resize transform to resize all training images to 384x768 without the two transforms above. That leads to an improved score (55.3 Q), which is the final score of our team. 
\end{itemize}

\subsection{\protect\msecond{} Mari-MobileSegNet}
\label{usv-eseg:marimobilesegnet}
\noindent
\emph{Yuan Feng, Lixin Tian}\\
\texttt{\small{\{fengyuan, 112023116tlx\}@dlmu.edu.cn
}}\\
\emph{Dalian Maritime University}\\
\textbf{Contributions:}  Conceptualization: YF; Implementation: LT; Experiments and Analysis: YF; Supervision: YF\\

\noindent\textbf{Method:} MobileNetV2~\cite{sandler2018mobilenetv2} is a novel mobile architecture that significantly improves the performance of mobile models across various tasks and model sizes. It introduces an inverted residual structure and lightweight depthwise convolutions to enhance model representation capabilities and reduce computational complexity. The approach emphasizes the importance of removing non-linearities in narrow layers and decoupling input/output domains from the expressiveness of the transformation. The submitted method utilizes a configuration file named \textit{mobilenet-v2-d8\_deeplabv3\_4xb4-160k\_ade20k-512x512} from the MMSegmentation~\cite{mmseg2020} code repository. In this configuration, the encoder utilizes MobileNetV2, while the decoder employs the ASPP decoding head from the DeeplabV3~\cite{chen2017rethinking} architecture.

\noindent\textbf{Backbone:} MobileNetV2~\cite{sandler2018mobilenetv2} pretrained on ImageNet-1k dataset with a resolution of 256x256

\noindent\textbf{Decode head:} ASPPHead~\cite{chen2017rethinking}

\noindent\textbf{Auxiliary head:} FCNHead~\cite{long2015fully}

\noindent\textbf{Training:}
\begin{itemize}
    \setlength{\itemsep}{1pt}
    \setlength{\parskip}{0pt}
    \setlength{\parsep}{0pt}
    \item BatchSize: During the training process, the batch size used was 4.
    \item Optimizer: SGD, learning rate: 0.01, momentum: 0.9, decay: 0.0005.
    \item Schedule: We adopted a warm-up strategy and the "poly" learning rate scheduler during training. The training process lasted for a total of 320,000 iterations, starting from iteration 1500 and ending at iteration 320,000.
    \item Augmentations: including random flipping, random cropping, random brightness adjustment, and random saturation adjustment, among others.
    \item Loss Functions: In this method, only the cross-entropy loss function is utilized.
\end{itemize}

\textbf{Datasets used:}
\begin{itemize}
    \setlength{\itemsep}{1pt}
    \setlength{\parskip}{0pt}
    \setlength{\parsep}{0pt}
    \item Backbone pretraining: ImageNet-1k
    \item Model pretraining: ADE20K 512X512
    \item Finetuning: LaRS train
\end{itemize}

\noindent\textbf{Hardware:} The training device is NVIDIA RTX 2080Ti with 11G memory

\noindent\textbf{Observations:} In order to reduce computational costs and enable deployment on embedded devices, we utilized a lightweight backbone network, MobileNetV2~\cite{sandler2018mobilenetv2}, which is consistent with the approach taken in eWaSR~\cite{tervsek2023ewasr}. The substitution of a more lightweight backbone network has effectively resolved computational cost issues, however, it may negatively impact detection performance, and strike a balance between detection performance and real-time processing is necessary when building a segmentation network. Additionally, during our attempts to export ONNX formats using other methods, we encountered errors caused by unsupported operators within the Pytorch-related library, such as global average pooling. Inspired by this, we explored the replacement of computationally intensive modules in the network with equivalent, simplified, and more efficient operators to also reduce computational costs when converting to ONNX formats.

\subsection{Baseline methods}
\label{usv-eseg:baselines}
\noindent
\emph{MaCVi Organizers}\\

\noindent
We train the SegFormer~\cite{xie2021segformer} with MiT-B2 backbone, DeepLabv3+~\cite{chen2017rethinking} and FCN~\cite{long2015fully} with ResNet-101~\cite{he2016deep} backbones, FCN~\cite{long2015fully} and BiSeNetv1~\cite{yu2018bisenet} with ResNet-50~\cite{he2016deep} backbones, STDC2~\cite{fan2021rethinking}, STDC1~\cite{fan2021rethinking}, and BiSeNetv2~\cite{yu2021bisenet} models on LaRS~\cite{Zust2023LaRS} dataset. We use MMSegmentation~\cite{mmseg2020} framework with default settings for training.


\section{USV-based Obstacle Detection}

\subsection{YOLOv8 Extra Large}
\label{usv-det:YOLOv8ExtraLarge}
\emph{Matej Fabijanić, Magdalena Šimunec, Nadir Kapetanović}
\texttt{\small{\{MatejFabijanic, MagdalenaSimunec\}@fer.hr}}\\
\emph{University of Zagreb, Faculty of Electrical Engineering and Computing}\\
\textbf{Contributions:} Conceptualization: MF, NK; Implementation: MF, MŠ; Experiments and Analysis: MF, MŠ, NK (analysis); Supervision: NK\\

The neural network architecture used for the challenge of obstacle detection was“YOLOv8 Extra Large” model~\cite{yolov8github} with 68.2 million parameters. We have not made any adaptations to the architecture, seeing how we decided to just test achievable detection accuracy for the architecture as-is before trying to make changes in order to enhance the model for the specific use-case of detecting objects on the surface of water. 

The architecture and its documentation can be found at~\cite{yolov8github}. We have used the default training parameters found at \cite{ultralytics_train_args}, with the exceptions of lowering the number of epochs to wait for no observable improvement for early stopping of training from 50 to 15 due to long training time, and lowering the number of images per batch from 16 to 8 due to experienced memory errors when using a higher batch size. We have not made any augmentations to the images.

\subsubsection*{Datasets Used for Pretraining}

Datasets used for pretraining [dataset name (train images/test images)]: Open Images Dataset V7 (15981/4019), COCO 2017 (621/155), CalTech256 + Boat Types Recognition dataset from Kaggle (376/93)\footnote{\url{https://mega.nz/file/UNFllTII\#} \url{fCood0Zd1QyfxQJOmvhycii2dx0mt-QZdanDaZhCr74}}, Šibenik Croatia 2023 CCTV footage (2034/299) \footnote{\url{https://mega.nz/file/wAk3zbyK\#YBz-njyY5v8QTs93aId5CtXDblPqkUmbO8xCAET_4fw}}.

For the Open Images dataset, we have used a randomly selected subset of 20000 images containing boats. For the COCO dataset, we have used a hand-picked dataset of 776 images containing boats which we deemed were useful for the purpose of the challenge. From CalTech256 + Boat Types Recognition dataset from Kaggle, we hand-picked and manually annotated 469 images of different boats, trying to have row boats and other classes also present.

\subsubsection*{Machine Specifications Used for Training}
CPU: Intel i9-9900K @3.60 GHz, GPU: NVIDIA GeForce GTX 1080Ti 11GB, RAM: 64 GB. \\
Inference Speed: 34ms for 2208x1242 pixels

\subsection{Co-DETR}
\label{usv-det:Co-DETR}
\emph{$^1$Andreas Michel, $^1$Wolfgang Gross, $^2$Martin Weinmann}
\texttt{\small{\{andreas.michel,wolfgang.gross\}@iosb.fraun\-hofer.de}, martin.weinmann@kit.edu}\\
\emph{$^1$Fraunhofer IOSB, $^2$Karlsruhe Institute of Technology}\\
\textbf{Contributions:} Conceptualization: AM; Implementation: AM,WG; Experiments and Analysis: AM,MW; Supervision: WG,MW\\

The used detection pipeline is based on the Co-DETR \cite{zong2023detrs} object detector. Co-DETR exploits multiple parallel auxiliary heads in order to increase the learning effectiveness of detection transformer. In our case, the underlying detector is DINO \cite{zhang2022dino} with six encoder- and decoder-layers with sinusoidal positional encoding. We used the implementation provided in the MMDetection toolbox version 3.2 \cite{chen2019mmdetection} to train our detector. As a backbone, we utilize an ImageNet-22K \cite{deng2009imagenet} pre-trained Swin-L \cite{liu2021swin} vision transformer.
Our training is conducted on two NVIDIA A100 80GB graphics cards.
The detector is pre-trained on the Objects365 \cite{shao2019objects365} and the MS COCO \cite{lin2014coco} datasets. In the next stage, we combine all images from MS COCO, including the category ship, with the LARS \cite{Zust2023LaRS} dataset and train for twelve epochs. The last training stage includes only the LARS dataset with the dynamic obstacle class and is only two epochs long. While training, we apply data augmentation in the form of random vertical image flipping, random multi-scale resizing, and random crop operations. AdamW \cite{loshchilov2018decoupled} was used as an optimizer with an initial learning rate of 0.0002 and a weight decay of 0.0001. Furthermore, we applied gradient clipping and apply a 0.1 learning rate multiplier to the backbone. 
Testing was performed on an NVIDIA A100 80GB graphics card. The inference speed was about four FPS. The test pipeline includes resizing to a 2048x1280 scale but no test time augmentation (TTA). In our experiments, TTA with multi-scale resizing leads not only to an expected large decrease of 3 FPS but also to an unexpected loss in detection performance. Furthermore, skipping the second training stage and training directly on the LARS dataset for twelve epochs decreases the F1 score by over three points on the test dataset. Varying the confidence score threshold also caused a large fluctuation in detection performance. The empirically determined score threshold of 0.25 results in the best F1 score for the chosen method.

\subsection{ScatYOLOv8+CBAM}

\label{usv-det:ScatYOLOv8CBAM}
\noindent
\emph{Borja Carrillo-Perez, Alexander Klein, Antje Alex, Edgardo Solano-Carrillo, Felix Sattler, Yannik Steiniger, Angel Bueno Rodriguez}
\texttt{\small{\{Borja.CarrilloPerez, Alexander.Klein, Antje.Alex, Edgardo.SolanoCarrillo, Yannik.Steiniger, Angel.Bueno\}\\@dlr.de}}\\
\emph{German Aerospace Center (DLR), Institute for the Protection of Maritime Infrastructures, Bremerhaven, Germany}\\
\textbf{Contributions:} Conceptualization: BCP, AK, AA, ESC, FS, YS, ABR; Implementation: BCP, AK; Experiments and Analysis: BCP, AK; Supervision: BCP, AK, AA, ESC, ABR \\

For the USV-based Obstacle Detection challenge, we chose the ScatYOLOv8+CBAM architecture described in\cite{carrillo23scatyolov8}. 
This object detector improves ship recognition in maritime contexts by synergizing their global and local features for a better perceptibility, with more prominent delineation (by scattering transform) and leveraging their location (by attention mechanism). 
We combined the detector with Slicing Aided Hyper Inference (SAHI)\cite{sahi23akyon} for data augmentation in training and inference for an improved small object detection.

The implementation of ScatYOLOv8+CBAM, compared to the standard YOLOv8\cite{yolov8github}, adds the ScatBlock, a 2D scattering-transform-based block at the beginning backbone and Convolutional Block Attention Modules (CBAM) to the prediction heads.

In the submission we adopted YOLOv8x, the largest YOLOv8, as it provided the best results during the pre-selection, and to which we added an extra prediction head to conform ScatYOLOv8x+CBAM+P. 

Prior to training, we used the procedure described in the SAHI method\cite{sahi23akyon} for data augmentation. 
We created the new dataset, based on LaRS dataset\cite{Zust2023LaRS}, with slices of size 640$\times$640 pixels with 20\% overlap. 
The augmented training and validation set contain 19913 and 1597 sliced images respectively. 

We trained the model, with random weight initialization, on the augmented dataset for 40 epochs using the default YOLOv8 training parameters.
The $F1$ score achieved by our method on the test set is 39.79. 
The SAHI parameters during inference include 640$\times$640 pixel slices with no overlap, Intersection over Union (IoU) as post-process match metric with 0.3 threshold, and a confidence score threshold of 0.1.
The training, validation and test of our approach use a single NVIDIA A100 GPU (CPU AMD EPYC 7713 64-Core Processor).

\section{USV-based Tracking}

\subsection{Detectors Ensemble}
\label{tr:UWIPLboatrack}
\emph{Hsiang-Wei Huang, Cheng-Yen Yang, Zhongyu Jiang, Sheng-Yao Kuan, Yuan-Hao Ho, Jenq-Neng Hwang}\\
\emph{University of Washington (UWIPL)}\\
\emph{\{hwhuang, cycyang, zyjiang, shengyao, yuanhh2, hwang\}@uw.edu}\\
\\
We use YOLOv8x coco-pretrained weight and further trained on Lars dataset (https://lojzezust.github.io/lars-dataset/) for 100 epochs with 0.001 learning rate, all the other training parameters follow the default setting of YOLOv8 repository (https://github.com/ultralytics/ultralytics). Since only objects that are boat class in the test set are annotated, only images of “boat” class from Lars dataset are used for training. We run BoTSORT on the test set following the default tracking parameter of Ultralytics BoTSORT implementation. We do not record the FPS but according to the original paper of BoTSORT, the FPS is roughly 6.6. The Lars-trained YOLOv8 + BoTSORT serves as our baseline and resulted in a HOTA of 0.191.

We found that the model trained on Lars dataset cannot accurately detect those boats when the boats are close to the USV, this is mainly because the boats in Lars dataset are relatively smaller, while some of the boats in the BoaTrack test set is much larger. To overcome this, we found that using COCO pre-trained weights can yield better results for those boats that are closer, so we implemented the Lars-trained YOLOv8x for all the test video except for sequence 75 after 3450 frames, when the boats in the video are closer and larger, we use the detection from COCO pre-trained YOLOv8x to conduct tracking. This detectors ensemble trick resulted in 0.215 in HOTA, which is the 1st place of the USV Multi-object tracking competition.

All the experiments are conducted on one V100.

\subsection{ReIDTracker Sea: BoaTrack \& SeaDronesSee}
\label{tr:reid_boatrack}
\emph{Kaer Huang, Aiguo Zheng, Weitu Chong, Kanokphan Lertniphonphan, Jun Xie, Feng Chen, Jian Li, Zhepeng Wang}\\
\emph{\{huangke1, zhengag, klertniphonp, xiejun, chenfeng13, lijian30, wangzpb\}@lenovo.com, wtzhong22@m.fudan.edu.cn}
\\
See Section \ref{tr:reidtracker-sea}.

\subsection{DLR-BoaTrack}
\label{tr:dlr_team}
\emph{Angel Bueno Rodriguez, Borja Carrillo-Perez, Alexander Klein, Antje Alex, Yannik Steiniger, Felix Sattler, Edgardo Solano-Carrillo}\\
\emph{{angel.bueno, borja.carrilloperez, alexander.klein, antje.alex, yannik.steiniger, felix.sattler, edgardo.solanocarrillo}@dlr.de}\\
\\

Our methodology employs the tracking-by-detection
paradigm: an object detection model identifies objects, and a
multiple object tracking algorithm links these detections into
coherent trajectories. All our experiments were performed on
an NVIDIA A100 GPU (CPU AMD EPYC 7713 64-Core
Processor). Detailed methodology is outlined below:
Object Detection: As object detector, we selected the ScatYOLOv8+CBAM architecture \cite{carrillo2023improving}. This object detector builds
upon the YOLOv8 framework to enhance ship recognition in
maritime environments, focusing on refining meaningful visual
features for accurate vessel detection and localization. To this
end, we adapt YOLOv8x and add an extra prediction head
to conform ScatYOLOv8x+CBAM+P. For pre-training data
augmentation, we utilized the SAHI workflow \cite{akyon2022slicing}, slicing the
LaRS dataset training and validation partitions into 640×640
pixel segments with a 20
slicing, the training and validation partitions of the LaRS
dataset produced an augmented dataset of 19913 sliced training images and 1597 sliced validation images. This augmented
dataset was used to train our ScatYOLOv8x+CBAM+P model
over 40 epochs, starting with random weights and the default
YOLOv8 training parameters. We excluded SAHI during inference to prioritize processing efficiency.
Tracking: The tracker employs the BYTE association mechanism \cite{bytetrack} with the camera motion compensation developed in
BoT-SORT \cite{aharon2022bot}. This amounts to using BoT-SORT with no re-identification module. The selection of confidence thresholds
for the first and second associations is performed empirically
by looking at the statistics of confidences generated by the
detector on each video. The default parameter configurations
of BoT-SORT were adjusted for video 366.avi, for which a
high-score threshold of 0.2, a threshold of 0.4 for new tracks,
and a threshold of 0.5 for confirming were used. For the rest
of the videos, the parameter configurations remain close to the
default values: high-score threshold of 0.5, new track threshold
of 0.6, and confirmation threshold of 0.7. After inference, the
results for all the videos are interpolated using the standard
procedure in the MOT challenge. Our method achieved a
HOTA of 0.193, MOTA of 0.057, and IDF1 of 0.202 on the
BoaTrack test set, running at an average of 12.6 FPS, including the object detection timings.

{\small
\bibliographystyle{ieee_fullname}
\bibliography{egbib}
}

\end{document}